\pdfoutput=1
\documentclass[11pt]{article}

\usepackage[]{emnlp2022}

\usepackage{times}
\usepackage{latexsym}

\usepackage[T1]{fontenc}
\usepackage[utf8]{inputenc}

\usepackage{microtype}
\usepackage{graphicx}
\usepackage{amsmath}
\usepackage{booktabs}
\usepackage{amssymb}
\usepackage{multirow}
\usepackage{float}
\usepackage{hyperref}
\usepackage{subfigure}
\usepackage{stfloats}
\usepackage{verbatim}
\usepackage{bbding}

\newcommand{\tabincell}[2]{\begin{tabular}{@{}#1@{}}#2\end{tabular}}

\def \dataset {META-GUI}
\title{\dataset: Towards Multi-modal Conversational Agents on Mobile GUI}

\author{Liangtai Sun\footnotemark[1], Xingyu Chen\footnotemark[1], Lu Chen\footnotemark[2], Tianle Dai, Zichen Zhu \and Kai Yu\footnotemark[2] \\
         X-LANCE Lab, Department of Computer Science and Engineering \\ MoE Key Lab of Artificial Intelligence, AI Institute, Shanghai Jiao Tong University \\ Shanghai Jiao Tong University, Shanghai, China \\
         \{slt19990817, galaxychen,  chenlusz\}@sjtu.edu.cn, \\
         \{daitl2000, jameszhuthethird,  kai.yu\}@sjtu.edu.cn}

\begin{document}
\maketitle

\renewcommand{\thefootnote}{\fnsymbol{footnote}}
\footnotetext[1]{Equal contributions.}
\footnotetext[2]{The corresponding authors are Lu Chen and Kai Yu.}

\begin{abstract}
  Task-oriented dialogue (TOD) systems have been widely used by mobile phone intelligent assistants to accomplish tasks such as calendar scheduling or hotel reservation. Current TOD systems usually focus on multi-turn text/speech interaction, then they would call back-end APIs designed for TODs to perform the task. However, this API-based architecture greatly limits the information-searching capability of intelligent assistants and may even lead to task failure if TOD-specific APIs are not available or the task is too complicated to be executed by the provided APIs. In this paper, we propose a new TOD architecture: GUI-based task-oriented dialogue system (GUI-TOD). A GUI-TOD system can directly perform GUI operations on real APPs and execute tasks without invoking TOD-specific backend APIs. Furthermore, we release \textbf{META-GUI}, a dataset for training a \textbf{M}ulti-modal conv\textbf{E}rsa\textbf{T}ional \textbf{A}gent on mobile \textbf{GUI}. We also propose a multi-model action prediction and response model, which show promising results on META-GUI. The dataset, codes and leaderboard are publicly available\footnotemark[3].
\end{abstract}
\footnotetext[3]{\url{https://x-lance.github.io/META-GUI-Leaderboard/}}

\section{Introduction}

\begin{figure}[t]
    \centering
    \includegraphics[width=0.47\textwidth]{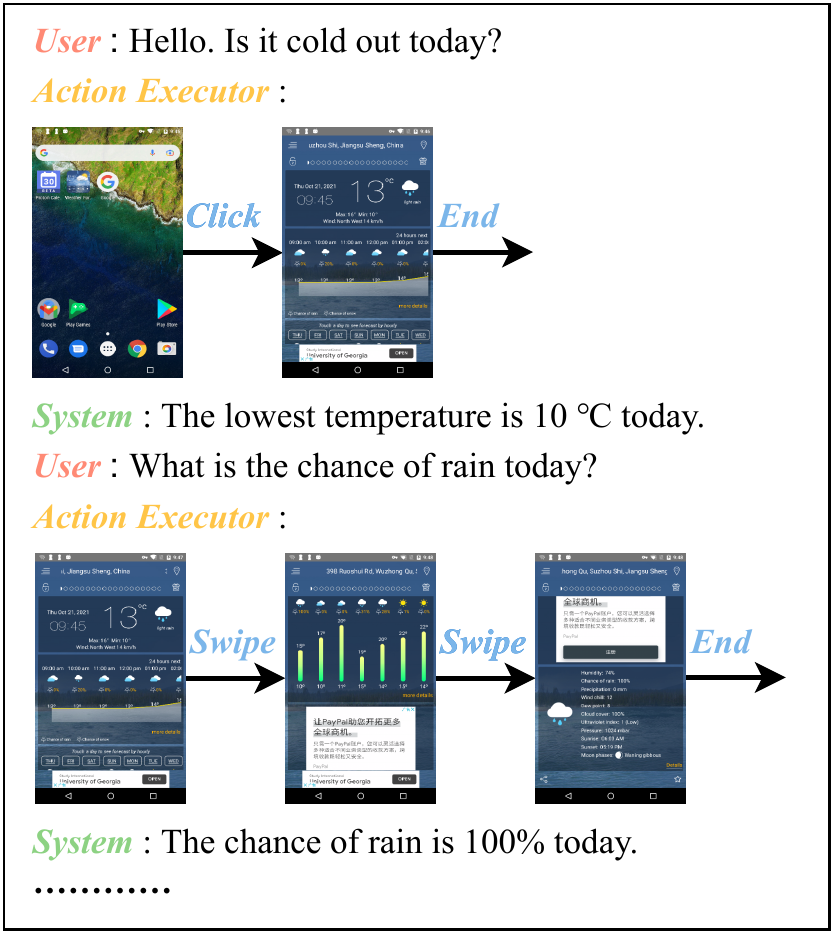}
    \caption{An example of the GUI-based task-oriented dialogue system(GUI-TOD). The Action Executor will execute tasks on GUI and the system will generate a response based on the execution result.}
    \label{fig:system_example}
\end{figure}

Recent years have witnessed the rapid development of task-oriented dialogue systems~\cite{zhang2020recent,ni2021recent,chen2022dialogzoo,chen2017survey}. They have been widely applied to customer support, booking system and especially intelligent personal assistant. These task-oriented dialogue systems work in a similar pipeline: firstly identify the user intent, then extract necessary information by the process of slot-filling. After getting enough information for the task, the agent will call the backend APIs (provided by APP developers) to fetch information, and then generate a response based on the query result.

There are some drawbacks of this framework. Firstly, TODs rely on publicly accessible APIs or APIs designed for TODs to perform tasks, but such APIs may not exist in real-life APPs, which hinders the application of TODs. Secondly, a system should be customized to recognize the pre-defined API-related slots, which limits the generality.

\begin{table*}[h]
\centering
\begin{tabular}{|l|l|l|}
\hline
Action & Description                                       \\ \hline\hline
\texttt{Click}(\texttt{item} = $x$) & Click the item with index $x$ on the screen. \\
\texttt{Swipe}(\texttt{direction} = $x$) & Swipe screen towards direction $x$, which includes ``up'' and ``down''. \\
\texttt{Input}(\texttt{text} = $x$) & Input the text $x$ to the smartphone.\\
\texttt{Enter}(\ ) & Press the ``Enter'' button on the keyboard. \\
\texttt{Clear}(\ ) & Clear the current input box. \\
\texttt{Back}(\ ) & Press the ``back'' button on the smartphone.    \\
\texttt{End}(\ ) & Turn has been finished and it will go to Response Generator module. \\ \hline
\end{tabular}
\caption{The actions in our dataset. There are 7 different actions with 3 different parameters.}
\label{action_table}
\end{table*}

Consider how humans perform tasks on smartphones They don't need a parametric API but finish tasks by interacting with the GUI (graphical user interface), indicating that GUI is a more general interface. 
%Considering the inconvenience of building agents on APIs, it's natural to incorporate GUI into task-oriented dialogue systems. 
Previous studies explore how to translate natural language commands into GUI operations~\cite{mazumder2021flin,pasupat2018mapping,xu2021grounding}. These studies focus on single query and step-by-step operations, while in real scenarios the query would be multi-turn interaction and there is no clear instruction about how to execute the task. Etan~\cite{riva2021etna} and SUGILITE~\cite{li2017sugilite} are two systems that support learning GUI operations from demonstrations, but these systems are script-based and are sensitive to the change in GUI and workflow. Duplex on the web~\cite{duplex} can directly operate the website to perform the required task, for example booking a movie ticket. However, it only supports limited websites, and it's more a unified GUI interface than a task-oriented dialogue system that enables general GUI operation.

To this end, we propose the task of GUI-based task-oriented dialogue system (GUI-TOD). It supports multi-turn conversation and direct GUI operation. All tasks would be performed on the GUI of real APPs, which means we no longer need TOD-specific APIs to communicate with APPs, and it would be possible to apply TOD on any APPs. Since there is no available benchmark published, We collect \dataset, a dataset with dialogues and GUI traces on real Android APPs. A GUI trace is a series of GUI operations, including screenshots, Android view hierarchies as well as actions. Android view hierarchy is an XML-style file, which organizes the content of GUI through a hierarchical structure. It also contains the types of items on the screen and their bounding boxes. An example is shown in Appendix \ref{view hierarchy}. When a user requests a task, the system should open the related APP and execute the task through multiple operations on GUI. It requires a comprehensive understanding of GUI structure and interaction logic. An interaction example is shown in Figure \ref{fig:system_example}.

%Our task is action prediction: given the GUI trace and dialogue history, predict the next action to be performed. Though GUI varies between different APPs, there is a common operation logistics behind the design. For example, clicking the "back" (or "←" icon) can  go back to the last screen, clicking the "+" icon will enter the procedure of adding a item.
 We focus on building an agent with general ability to operate GUI, rather than optimize for specific APPs.
 Our proposed GUI-TOD system leverages both the visual information and textual information on the screen to predict the next action to be executed and generate the system response. Our experiments show that the GUI-TOD outperforms heuristic baselines by a large margin, with an action completion rate of 82.74\%.

Our contributions are followings:
\begin{itemize}
    \item We propose a GUI-based task-oriented dialogue system, which can perform tasks on mobile APPs through multiple operations on GUI.
    \item We collect \dataset{}, a dataset with dialogues and GUI operation traces serving as the benchmark for the proposed system.
    \item We conduct thorough experiments on our dataset and validate the importance of multi-modal information and history information. We show that it is a promising task but needs further exploration.
\end{itemize}

\section{Task Definition}
\label{taskdefination}

\begin{figure}[h]
    \centering
    \includegraphics[width=0.47\textwidth]{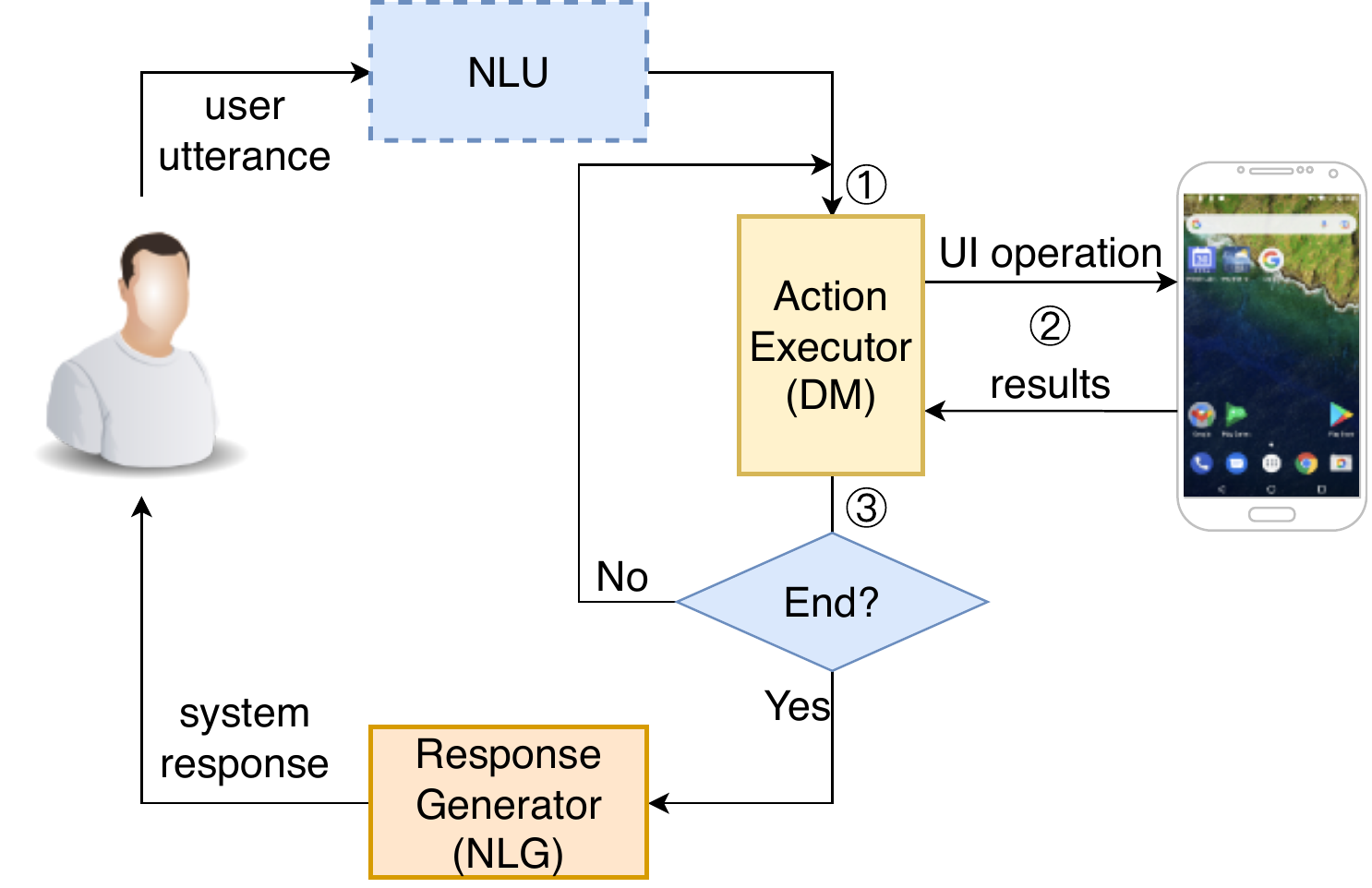}
    \caption{The overview of GUI-based task-oriented dialogue system (GUI-TOD).}
    \label{fig:gui-tod}
\end{figure}

The overview of GUI-TOD is shown in Figure \ref{fig:gui-tod}. It consists of two sub-modules: Action Executor (AE) and Response Generator (RG).
The traditional task-oriented dialogue system~\cite{chen2017survey,zhang2020recent,yu2014cognitive} splits the task into natural language understanding (NLU) \cite{zhu2021few}, dialogue manager (DM) \cite{chen2020schema,zhu-etal-2020-efficient,chen2018policy,chen2019agentgraph,chen2020distributed}, and natural language generation (NLG) \cite{keskar2019ctrl}. We omit the NLU module and directly send user utterances to AE. The AE module has similar features with DM, it executes the requested task by interacting with the GUI for multiple rounds, while DM accomplishes this by calling TOD-specific APIs. The RG module will generate the system response based on the execution results, which is the same as NLG. The process of executing a task is a series of GUI operations, including click, swipe, etc. The task of AE module is action prediction, which aims at predicting the next action to be performed on GUI, and the RG module focuses on generating system's response after executing a task. A major improvement of GUI-TOD is that it does not rely on a pre-defined domain ontology. Conventionally, the DM module will identify a set of slot-value from the user utterance, which serves as the parameter for backend APIs. However, GUI-TOD handles task-specific slot-values during the execution of tasks. When the APP requires a certain input (for example, entering the time and destination), the system can obtain the information by understanding the current user utterance or generating a response for further asking. 
%The slot-filling-like procedure will only happen when the system encounters the input request from APPs. 
Compared with CUED actions \cite{young2007cued} in traditional TOD, actions in GUI-TOD are GUI-related operations rather than communication actions between user and system.

Formally, the action prediction task can be defined as: given the GUI trace and dialogue history, predict the next action to be performed. 
We define the set of actions that can be performed on the APPs in Table \ref{action_table}. All the actions would take the form of $Action(parameter=*)$. There are seven types of $Action$, including six physical actions: \textit{click, swipe, input, enter, clear, back}, and one virtual action: \textit{end}. The corresponding parameters are listed in Table \ref{action_table}. The $end$ action is the last action for every GUI trace, which means the end of GUI operations. After an $end$ action is generated, the GUI-TOD would move to the RG module.
We denote the \textit{j}th action in turn \textit{i} as $\mathcal{A}_{i,j}=\left( t, p \right)$, where \textit{t} is the action type and \textit{p} is the corresponding parameter. $\mathcal{S}_{i,j}=\left( s, v \right)$ is the \textit{j}th screen in turn \textit{i}, including the screenshot \textit{s} and the view hierarchy \textit{v}. The dialogue in turn \textit{i} is represented as $\mathcal{D}_{i}=\left( U_i, R_i \right)$ where $U_i$ is the $i$th user utterance and $R_i$ is the $i$th system response. The action prediction task is formulated as:
\begin{equation}
    \mathcal{A}_{i,j}=\mathcal{F}\left( \mathcal{S}_{1:i, 1:j}, \mathcal{A}_{1:i,1:j-1}, \mathcal{D}_{1:i-1}, U_i \right),
\end{equation}
where $1:i$ means from turn $1$ to $i$, $\mathcal{F}$ is a trainable action model, which we discuss in \ref{actionmodel}. The RG module takes the GUI trace and dialogue history as input, then generates a response based on the execution result and context. Denote the set of actions in turn $i$ as $\mathcal{A}_i$, the screens in turn $i$ as $\mathcal{S}_i$, the response generation task is formulated as:
\begin{equation}
    \mathcal{R}_{i}=\mathcal{G}\left( \mathcal{S}_{1:i}, \mathcal{A}_{1:i}, \mathcal{D}_{1:i-1}, U_i \right),
\end{equation}
where $\mathcal{G}$ is the response generator model, which we discuss in \ref{responsemodel}. 
%The response could be a simple confirmation like "is this good?" or a description for the current screen, for instance describing the weather shown on the screen.

\section{Meta-GUI Creation}
Our dataset consists of two kinds of data: dialogues and GUI operation traces. In each dialogue, user would ask the agent to complete a certain task through multi-turn interaction. Our tasks involve six different domains: weather, calendar, search, taxi, hotel and restaurant. 
%The agent should complete the task by performing GUI operations on real APPs. 
In this paper, we consider APPs that accomplish the same kind of tasks to be in the same domain.
To enhance the diversity of our dataset, we use multiple Apps from the calendar and weather domains.
%All the APPs we used are downloaded from GooglePlay, 
The details of APPs are listed in Appendix \ref{app_data}.

\begin{figure*}[h]
    \centering
    \includegraphics[width=1\textwidth]{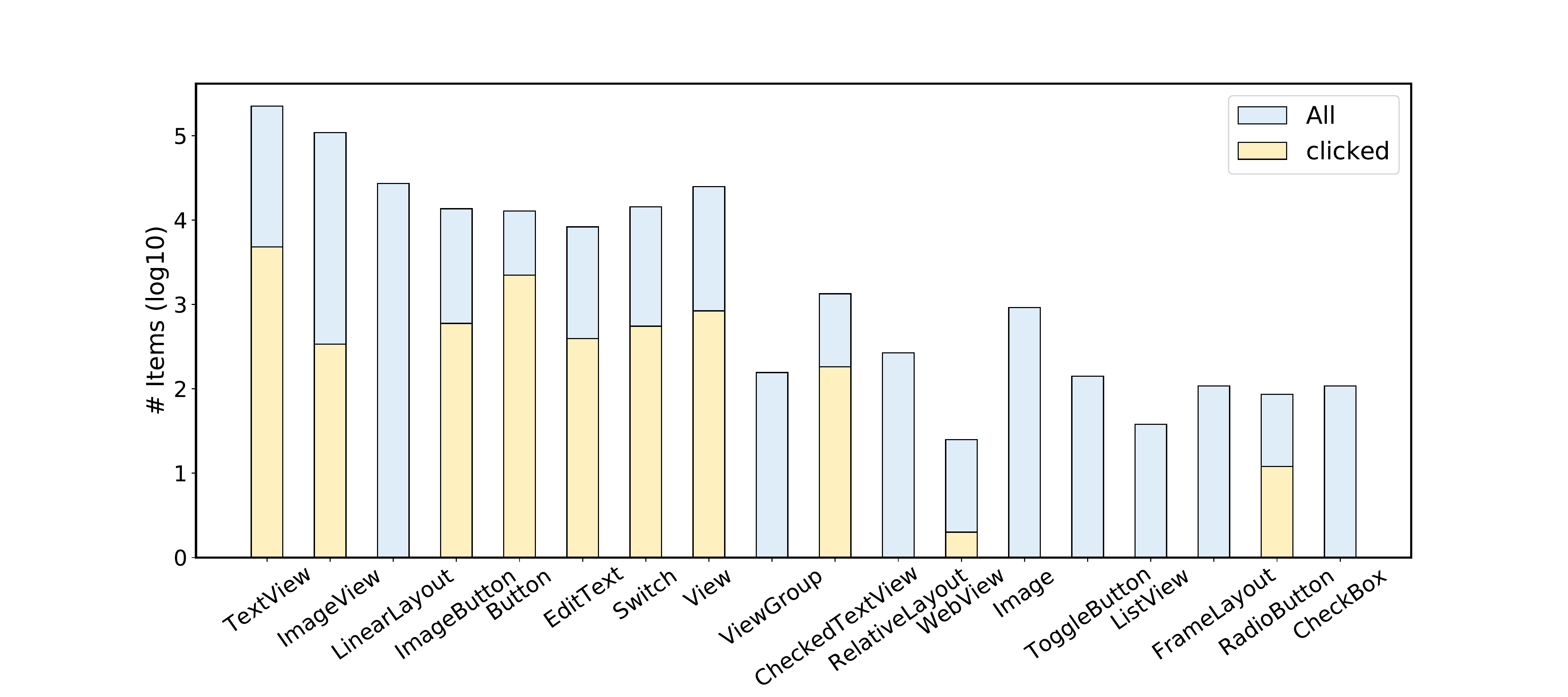}
    \caption{The distribution of the total number of items versus the clicked one for each item type.}
    \label{fig:item_type}
\end{figure*}

\subsection{Collecting GUI traces}
We collected our data in two-stage: first we collected GUI traces for existing dialogues, then we collected both dialogues and GUI traces.% for informative APPs.

In the first stage, we provided dialogues to annotators and instructed them to perform tasks on real APPs. We started from extracting dialogues from the SMCalFlow dataset \cite{andreas2020task}. SMCalFlow contains multi-turn task-oriented dialogues, which is known for complex reference phenomenon that requires a comprehensive understanding of context. 
We extract dialogues from calendar, weather and search domains. Six annotators were recruited to label the GUI traces. We built a web-based annotation system, which was connected to a real Android smartphone (see Appendix \ref{annotation system}). Annotators can see the current screen of the smartphone in the system, and control the smartphone by clicking buttons. A dialogue would be shown in the system. 
Annotators should first read the dialogue, then they were allowed to explore how to finish the task (e.g. check the weather) on smartphone. 
If the task requirement in the dialogue conflicted with the real-world scenario (for example, creating an event in the past), the annotators could change the content of the dialogue to make the task achievable. 
After they were ready, they need to use the annotation system to record the actual process of executing the task. Each operation would be recorded, and the screenshot after each operation was also saved together with the view hierarchy.
%In this figure, the left part shows a screenshot of the Google Calendar App, and part of the view hierarchy is shown on the right. The selected view is also highlighted on the left.

In the second stage, we collected dialogues and GUI traces for domains of hotel, restaurant and taxi. Because there are no available dialogues of these domains in previous datasets, we asked annotators to write new dialogues. We selected three experienced annotators from the last stage. Different from the last stage, the annotator was shown a task objective, which was generated randomly from all available conditions in APPs. The annotators should act as user and system alternatively to write dialogues according to the task objectives. To avoid annotators writing short and simple dialogues, we added constraints about the number of turns and the behaviors in dialogue, e.g. adding a condition or changing a condition. An example of the generated target is shown in Appendix \ref{exampleoftarget}. After writing dialogues, the annotators should also record the corresponding GUI operation traces for each turn, which is the same as the last stage.

\subsection{Data Review}
After annotation, we manually reviewed the data. The checklist includes: whether the recorded GUI traces match the dialogues, whether there are invalid operations due to the system error or misoperation, and whether there are redundant operations in the GUI trace. We manually fixed annotations that only have small mistakes, and discarded the task requiring significant modification. The dialogue level pass rate is about $63.6\%$, and finally we got 1125 dialogues in total. For more information, please refer to Appendix \ref{datareview}.

\subsection{Post-processing}

The dialogues collected in the second state were created by three annotators, which lack diversity in expression. Therefore, we published a dialog rewritten task on AMT\footnote{\url{https://www.mturk.com/}} (Amazon Mechanical Turk) to polish the dialogues. 
%We only select utterances with more than 15 words.

\begin{figure}[h]
    \centering
    \includegraphics[width=0.5\textwidth]{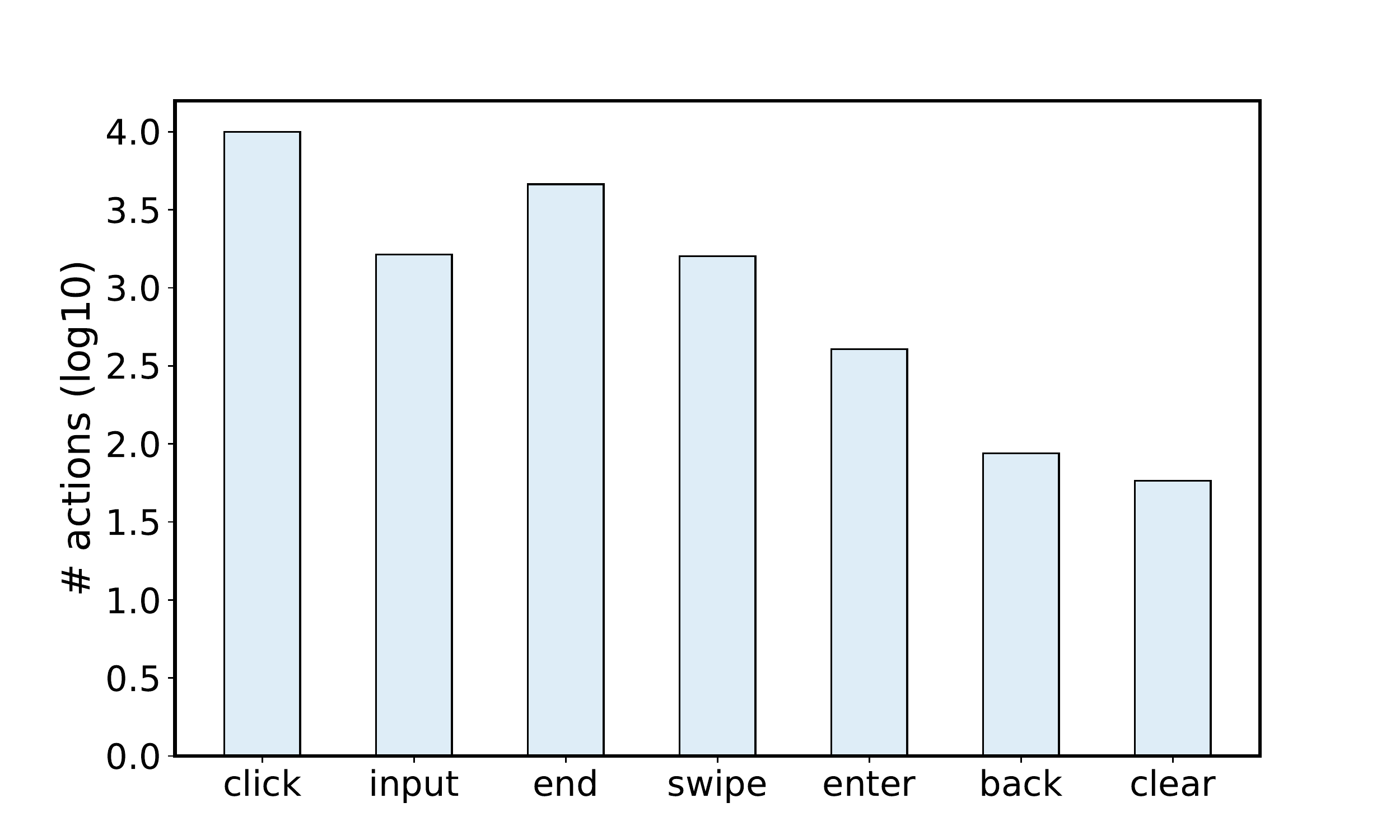}
    \caption{The distribution of actions.}
    \label{fig:action_distribution}
\end{figure}

During GUI trace annotation, some APPs can not obtain valid Android hierarchy. To handle this problem, we used the online Optical Character Recognition (OCR) service, provided by Baidu Cloud \footnote{\url{https://cloud.baidu.com/}}, to detect all texts on the image with their corresponding positions and generate a pseudo layout file.

\begin{figure*}[h]
    \centering
    \includegraphics[width=1.0\textwidth]{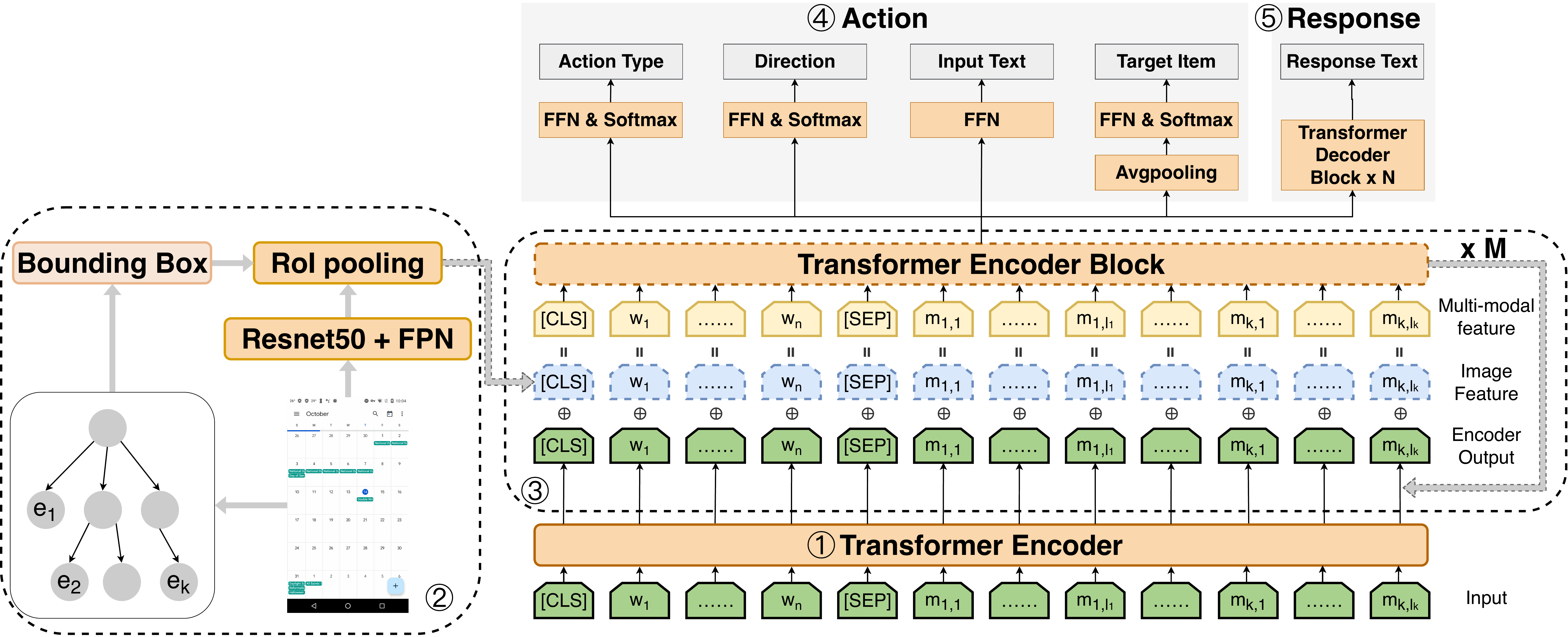}
    \caption{The illustration of our proposed model. There are five parts in this figure: (1) encoder; (2) image feature extraction; (3) multi-modal information fusion; (4) the Action Module; (5) the Response Module.}
    \label{fig:armodel}
\end{figure*}

We extract items from screen using the corresponding layout file. An item is a clickable leaf node. Similar to \cite{zhou2021large}, we consider an item to be clickable if its \texttt{clickable} attribute is true or its parent node is clickable. An item consists of text content, item type and bounding box. 
We extract the text content of an item by looking at its \texttt{text} property first. If it is empty, we use its \texttt{content-desc} attribute, otherwise we would use the \texttt{resource-id} property. Based on the extracted items, we can locate the target item for the \texttt{click} action by comparing the click position and the bounding boxes of items. 

\subsection{Data Analysis}

The total number of dialogues in our dataset is 1125, including 4684 turns. %(a turn is a pair of user and system utterance). 
The average number of images for each turn is 5.30, and the average number of words for each utterance is 8. On average, there are 23.80 items for each image, and the item text length is 2.48 words. The distribution of item types is shown in Figure \ref{fig:item_type}. We also provide an example for each item type in Appendix \ref{exampleofitemtypes}. It is clear that \texttt{TextView} and \texttt{ImageView} are the two most frequent type, which indicates that our dataset is informative.
%The mean length of user utterance is 6.8 tokens and 9.5 for system utterance. We display a sunburst graph of utterance in Figure ?. The most frequent sentence is "?". 

 The distribution of actions is listed in Figure \ref{fig:action_distribution}. The \texttt{click} is the most frequent action, while \texttt{clear} is the least action for the reason that only a small number of tasks require clearing the current input box. For \texttt{click} action, we further compute the type distribution of target items, which is shown in Figure \ref{fig:item_type}. \texttt{TextView} and \texttt{Button} type are mostly clicked, while there are 8 item types never been operated. This implies that the item types may supply some hints for predicting the target items. Besides, the average numbers of words for \texttt{response} and \texttt{input} action are 9 and 3 respectively.
 %The distribution of GUI trace length on different tasks is shown in Figure ?. 

% app analysis
% We used 11 different APPs for six tasks, the data distribution on APPs is shown in Appendix \ref{app_data}. %todo todo

%\section{GUI Based Task Oriented Dialogue System}
\section{Model Design}

The overview of our system is illustrated in Figure \ref{fig:armodel}. It's composed of four components: encoder, image feature extractor, multi-modal information fusion module and the output module. The output module can be the Action Module or the Response Module.

\subsection{Action Model}
\label{actionmodel}
%Thence, We apply tokenizer on the input text and the dialog history to extract the start position and end position at the token level. 
% The simplest Action Model only consists of encoder and action module. To verify the importance of multi-modal information, we further add it to the model, and get the multi-modal Action Model, which consists of these four parts. Next, we will introduce these separately.
We call the combination of encoder, image feature extractor, multi-modal information fusion module and the Action Module as Action Model, which is used to predict the next GUI action based on the history. Next, we will describe these modules respectively. For simplify, for the screen history we only consider the last screen here. We will discuss adding more screen histories later.

\paragraph{Encoder} The input of encoder consists of two parts: dialog history $\{\mathcal{D}_{1:i-1}, U_i\}=\{w_{1},...,w_n\}$ and texts in the items $\{m_{1,1:l_1}, \dots, m_{k,1:l_k}\}$. Items are extracted from the last screen, $k$ is the number of items and $l_i$ is the length of the $i$th item's text:
\begin{align}
\begin{split}
    X &= \{ w_{1:n};m_{1,1:l_1}, \dots, m_{k,1:l_k}\}, \\
    \textbf{H} &= \text{TransformerEncoder}(X), \\
\end{split}
\end{align}
where $\textbf{H} = \left[\textbf{D};\textbf{M}\right]$ and $\textbf{D}=\{ \textbf{w}_1, \textbf{w}_2, \dots, \textbf{w}_n\}$ represents encoder outputs of the dialogue history, $\textbf{M}=\{ \textbf{m}_{1,1:l_1}; \dots; \textbf{m}_{k,1:l_k} \}$ represents encoder outputs of item texts.

\paragraph{Image feature extractor} Given a screenshot and its corresponding layout file, we use Faster R-CNN~\cite{ren2015faster} to extract the feature map. Then we apply ROI pooling based on the bounding box of each item, and get the item-level image features $\mathbf{I}=\{\mathbf{I}_1, ..., \mathbf{I}_k\}$. 

\paragraph{Multi-modal information fusion module} Given the encoder output and the regional image feature extracted above, we concatenate them together. The text features from one item $\textbf{m}_{i,1:l_k}$ are concatenated with the same item feature $\mathbf{I}_i$, and the $\mathbf{w}_{1:n}$ are concatenated with zeros. Then we use a Transformer encoder with $M$ layers to fuse the multi-modal features. For each layer, to enhance the image information, we will concatenate the image features and the output from the last layer again to form the input for the next layer.

\paragraph{Action Module} For the Action model, we need to predict the action type and its corresponding parameters. As shown in Table \ref{action_table}, there are 7 action types with 3 different parameters. We show some examples of parameter predictions in Appendix \ref{parameterpredictions}. 

We use the encoder output of the \texttt{[CLS]} token for action type prediction. We apply a feed-forward network followed by a Softmax layer to predict the action type:
\begin{align}
\begin{split}
    \mathbf{p}_a &= \text{Softmax}(\text{FFN}_1(\mathbf{E_{[CLS]}})),
\end{split}
\end{align}
where $\mathbf{p}_{a}$ is the probability distribution of action, and FFN represents the Feed-Forward Network.

For the action parameter, we use three different classifiers:

1) \textit{Input Text Prediction} \ We assume that the input to the APPs must be part of the user utterance, so we formulate the prediction of input text as a span prediction task. We use two classifiers to predict the begin and end positions in the dialogue:
\begin{equation}
    \begin{split}
        \mathbf{p}_{ds}=\text{FFN}_2(\mathbf{D}),
        \mathbf{p}_{de}=\text{FFN}_3(\mathbf{D}),\\
   \end{split}
\end{equation}
where the $\mathbf{p}_{ds}$ and $\mathbf{p}_{ds}$ are the probability of start and end position respectively. 

2) \textit{Target Item Prediction} \ The target item classifier is based on the encoding outputs of items. We first computed the item representation by applying average pooling on the encoding outputs, then we use a feed-forward layer to compute the probability of selecting an item followed by a Softmax layer:
\begin{equation}
    \begin{split}
    \bar{\textbf{m}}_i =& \  \text{Avgpooling}(\textbf{m}_{i,1:l_i}) \ \ 1 \leq i \leq k, \\
    \bar{\textbf{m}} =& \left[\bar{\textbf{m}}_1, \dots, \bar{\textbf{m}}_k\right]. \\
    \mathbf{p}_{m} =& \ \text{Softmax}(\text{FFN}_4(\bar{\textbf{m}})),
    \end{split}
\end{equation}
where $\mathbf{p}_{m}$ is the probability distribution of items.

3) \textit{Direction Prediction} \ The direction classifier is a two-classes classification layer for the direction \textit{up} and \textit{down}:
\begin{equation}
    \mathbf{p}_d = \text{Softmax}(\text{FFN}_5(\mathbf{E_{[CLS]}})),
\end{equation}
where $\mathbf{p}_{d}$ is the probability distribution of swipe direction.

\paragraph{Adding history information} According to the task definition, besides dialogue histories, we can still use action histories and screen histories. To verify this, we add them to the action model. For action histories, we regard action types as special tokens and add them to the dictionary. We concatenate the most recent $H$ action types $\{t_{1:H}\}$ before the dialogue history as input:
\begin{equation}
    X = \{t_{1:H};w_{1:n};m_{1,1:l_1},\dots, m_{k,1:l_k}\},
\end{equation}
where $X$ stands for the input of Encoder, $t$ represents the action type. 

For screenshot histories, we encode all the screenshot in a recurrent way. Assume $\hat{\mathbf{I}}_i=\left[\mathbf{I}_{i,1},...,\mathbf{I}_{i,k}\right]$ is the image feature for $i$th screenshot, and $\bar{\mathbf{I}}_i$ is the history image feature for time step $i$. We compute $\bar{\mathbf{I}}_{i+1}$ by:
\begin{equation}
\begin{split}
    \bar{\mathbf{I}}_{i+1} = \text{Attn}(\mathbf{W}_1&\hat{\mathbf{I}}_{i+1},\mathbf{W}_2\bar{\mathbf{I}}_{i},\mathbf{W}_3\bar{\mathbf{I}}_{i}), \\
    &1 \leq i \leq H-1,
\end{split}
\end{equation}
where $\bar{\mathbf{I}}_{1} = \hat{\mathbf{I}}_{1}$, $H$ is the length of history, $\text{Attn}$ is the attention mechanism~\cite{vaswani2017attention}, and $\mathbf{W}_*$ are trainable parameters. We use the $\bar{\mathbf{I}}_{H}$ to replace the image features in Figure \ref{fig:armodel}.

\subsection{Response Model}
\label{responsemodel}

The Response Model aims to generate the response to user. We use the Response Module as the output module and the other parts are the same as Action Model. Considering the prediction of response is mainly decided by the execution results and dialogues, we do not use action histories for the Response Model. For the Response Module, we use a Transformer Decoder with N layers:
\begin{align}
\begin{split}
    \textbf{R} = \text{TransformerDecoder}(\left[\textbf{D};\textbf{M}\right]),
\end{split}
\end{align}
where $\textbf{R}$ represents the predicted response text.

% The reply model is similar to the action model, except that we do not input the items to the model, shown in Fig. \ref{fig:replymodel}, since we believe that using the embedding methods of items like that of Action model is low-efficient. However, we still collect the text attributes of all items to get the input of page text.

% We use Vision Transformer \cite{dosovitskiy2020image} Embedding to extract the features of screenshot and use BERT\cite{devlin2018bert} Embedding to extract the features of dialog and page text. We concatenate them all and then feed them into a N-layer transformer encoder. For the decoder part, we use a M-layer transformer decoder.

\section{Experiment}
\subsection{Data Preprocess}
\label{dataset_generateion}

\begin{table}[h!]
\centering
\begin{tabular}{|c|c|c|c|}
\hline
                & Train & Dev  & Test \\ \hline
\# dialogues & 897   & 112  & 116  \\ \hline
\# turns     & 3692  & 509  & 483  \\ \hline
\# data      & 14539 & 1875 & 1923 \\ \hline
\end{tabular}
\caption{Dataset Statistics}
\label{tab:dataset_statistics}
\end{table}

We process the dataset in the granularity of action. Each data point takes as input the screenshot history, action history, dialogue history and predicts the action to be performed. 
%Due to the complexity of the view hierarchy and inaccuracy of OCR results, some click actions may lack target items and we will discard these data points. As for the \texttt{input} action, we claim that all the texts input is a sub-string of the user utterance of the current turn, so we use the start and end position in the utterance to represent an input text.
We obtained 18337 data points in total, and we randomly divide the data into the training set, development set and test set with the ratio of 8:1:1. The data statistics are shown in Table \ref{tab:dataset_statistics}.

\begin{table*}[thp!]
\centering
\resizebox{\textwidth}{35mm}{
\begin{tabular}{l|ccc|cccccc|c}
\toprule
\multirow{2}{*}{Method} & \multicolumn{3}{c|}{Information} & \multicolumn{6}{c|}{Action} & 
\multirow{2}{*}{\begin{tabular}[c]{@{}c@{}}Turn\\ CR\end{tabular}} \\
\cline{2-10} & mm & act\_h & scr\_h & \tabincell{c}{Action \\ Type Acc.} & \tabincell{c}{Input \\ EM} & \tabincell{c}{Input \\ F1} & \tabincell{c}{Item \\ Acc.} & \tabincell{c}{Direction \\ Acc.} & CR &  \\
\midrule
Random     & & &         & 14.02      & 8.72   & 17.96   & 9.08    & 51.26         & 5.37                                                      & 3.99                                                    \\
MFM & & & & 53.71      & 14.02   & 37.78   & 16.58    & 89.31         & 8.91                                                      & 0.00                                                    \\
FM & & &    & 37.48      & 6.65   & 14.02   & 9.94    & 81.51         & 10.00                                                      & 6.76                                                    \\
\midrule
LayoutLMv2 & \Checkmark & & & 85.60 & 47.37 & 70.76 & 64.38 & 92.95 & 64.48 & 36.88 \\
LayoutLM      & & & & 82.22      & 83.04   & 90.56   & 71.98    & 94.87        & 67.76  & 38.12  \\
BERT      & & &     & 87.52      & \textbf{93.57}   & \textbf{97.24}   & 82.84    & 93.59         & 78.42                                                      & 52.08                                                    \\
\ \ \ \ \ \ \ \ \ \ +mm & \Checkmark & & & 88.35 & 92.98   & 96.42   & 84.51    & 94.23 & 80.45 & 53.96 \\
\ \ \ \ \ \ \ \ \ \ +act\_h & & \Checkmark & & 88.87   & 91.81   & 94.86   & 84.23    & \textbf{95.51} & 80.97  & 55.42 \\
\ \ \ \ \ \ \ \ \ \ +scr\_h & \Checkmark & & \Checkmark & 89.86 & 90.06   & 95.30   & 84.32    & 94.87 & 81.54  & 55.62 \\
% BERT(mm)+act\_h  & 0.9007 & \textbf{0.9415}   & 0.9638   & \textbf{0.8673}    & \textbf{0.9551} & \textbf{82.84}  & 56.46 \\
%BERT(mm)+all\_h  & \textbf{0.9080} & 0.9123   & 0.9642   & \textbf{0.8590}    & 0.9423 & \textbf{82.74}  & \textbf{56.88} \\
m-BASH & \Checkmark & \Checkmark & \Checkmark & \textbf{90.80} & 91.23   & 96.42   & \textbf{85.90}    & 94.23 & \textbf{82.74}  & \textbf{56.88} \\
\bottomrule
\end{tabular}}
\caption{The experiment results of the Action Model on the test set. \textbf{Acc.}: accuracy. \textbf{EM}: Exact Match. \textbf{F1}: F1 score. \textbf{CR}: completion rate. \textbf{MFM}: Most Frequent Method. \textbf{FM}: Frequency Method. \textbf{mm}: use the multi-modal information fusion module to add image information. \textbf{act\_h}: add action histories. \textbf{scr\_h}: add screenshot histories.}
\label{tab:action_model_results}
\end{table*}

\subsection{Experiment Setup}
We train our baselines on the training set and select the best models on the dev set based on the Action completion rate. 
We use pretrained BERT~\cite{devlin2018bert}, LayoutLM~\cite{xu2020layoutlm} and LayoutLMv2~\cite{xu2020layoutlmv2} as our encoder models.
\footnote{There are some pre-trained models about GUI understanding, like ActionBERT~\cite{he2020actionbert} and UIBERT~\cite{bai2021uibert}. But they are not open-source.} 
BERT is pretrained on pure text corpus by masked languages modeling task, while LayoutLM and LayoutLMv2 are pretrained on scanned documents by masked visual-language modeling task and incorporate image features.
% We use three different pretrained models as our encoder:
% \begin{itemize}
%     \item \textbf{BERT}~\cite{devlin2018bert} \ BERT is a pre-trained bidirectional language model. BERT is pre-trained with two tasks, Masked Language Model (MLM) and Next Sentence Prediction (NSP). It has shown great success in many NLP tasks. 
%     \item \textbf{LayoutLM}~\cite{xu2020layoutlm} \ LayoutLM is a variant of BERT. It jointly models interactions between text and layout information. It also leverage image features to incorporate visual information. LayoutLM achieves great results in several VDU tasks.
%     \item \textbf{LayoutLMv2}~\cite{xu2020layoutlmv2} \ LayoutLMv2 models the interaction among text, layout and image in a single multi-modal framework. It uses not only the existing masked visual-language modeling task but also the new text-image alignment and text-image matching tasks. Experiment results show that LayoutLMv2 outperforms LayoutLM by a large margin.
% \end{itemize}

We use a batch size of 4 and fine-tune for 8 epochs. We use Adam optimizer with the learning rate of 1e-5. For Response Model, the number of Transformer Decoder Block is 4. Furthermore, we use three heuristic methods in our experiments:

\textbf{Random} We randomly predict action type and its corresponding parameters.

\textbf{Frequency Method (FM)} We first calculate the frequency of each action type and its corresponding parameters. Then, we apply the results to the development set and generate the prediction according to the frequency.

\textbf{Most Frequent Method (MFM)} Similar to the frequency method, we generate the prediction with the most frequent result.

For the evaluation, we use completion rate for action prediction. We first define two completion rate metrics: action completion rate and turn completion rate. One action is regarded as completed only if the action type and its parameters are correctly predicted. And if all actions in the same turn are completed, then the corresponding turn will be considered completed. For action type prediction, item prediction and direction prediction, we use accuracy. For input prediction, we use token level exact match and F1. 
And we use BLEU score to evaluate the Response Model.

\subsection{Experiment Result}
The experiment results of the Action Model are listed in Table \ref{tab:action_model_results}. 
We can find that the deep learning methods outperform the heuristic methods by a large margin, which is expected.
Comparing the results of BERT backbone and LayoutLM backbone, we find that BERT model yields better performance. The reason is that LayoutLM model was pre-trained on a scanned document image dataset, and there exists a large gap between the Android GUI and the scanned document images. Furthermore, we can find that LayoutLMv2 performs worse than LayoutLM. We hypothesize that LayoutLMv2 uses early-fusion method, which will bring more noises. We can also find that adding multi-modal information to BERT leads to a better performance ($52.08\% \rightarrow 53.96\%$), and the improvements are mainly from the action type prediction, target item prediction and swipe direction prediction. The reason why adding images would help is that the image information contains some action histories that cannot be represented by text. For example, when filtering conditions on hotel reservations, the conditions selected in the previous action can be seen through the image (as a highlighted text), but they can not be reflected through text. An example is illustrated in Appendix \ref{casestudy}. 
Besides, the image information can help the model to locate the item more accurately. For example, for a screen with multiple radio buttons, since the BERT model does not take the item position as input, the model cannot distinguish the corresponding button for each option by only textual input. 
However, we also find that the performance of input text prediction degrades after adding image information. We assume that BERT itself can successfully model text information, but adding visual information will affect the model's ability to understand text.

We further verify the importance of history information by adding action histories and screenshot histories. From the experiment results, we find that adding history information to BERT can improve the performance
($52.08\% \rightarrow 55.42\%$ after adding action history to BERT, $53.96\% \rightarrow 55.62\%$ after adding screenshot history to BERT+mm).
Adding action histories leads to greater performance improvement, which means action sequence is a more effective way to represent history. The screenshots contain higher-level history information, but the screenshot changes a lot before and after operation (sometimes one click may change the screen completely), which will bring difficulties to the information fusion.

Finally, we add all information, including multi-modal information, action histories and screenshot histories, to the BERT model and get the m-BASH (\textbf{m}ulti-modal \textbf{B}ERT with \textbf{A}ction histories and \textbf{S}creenshot \textbf{H}istories), which results in the state-of-the-art performance ($56.88\%$).

The results of the Response Model are shown in Table \ref{tab:response_model_results}. BERT outperforms LayoutLM and LayoutLMv2 by a large margin, which is consistent with the results of Action Model. We also find that adding multi-modal information and screenshot histories can improve performance, which means the model leverage the information from history to generate response.

\begin{table}[]
\centering
\begin{tabular}{lc}
\toprule
Method              & Response BLEU score \\
\midrule
Random              & 0.0071         \\
MFM & 0.0929         \\
FM     & 0.0788         \\
\midrule
LayoutLM           & 0.5043         \\
LayoutLMv2 & 0.5820 \\
BERT             & 0.6219 \\
\ \ \ \ \ \ \ \ \ \ +mm & 0.6224 \\
\ \ \ \ \ \ \ \ \ \ +scr\_h & \textbf{0.6311} \\
\bottomrule
\end{tabular}

\caption{The experiment results of Response BLEU score on the test set.}
\label{tab:response_model_results}
\end{table}

\subsection{Generality}
According to the design of our system, it does not need to pre-define API-related slots, therefore our system has a strong generality and can be easily adapted to new APPs. To demonstrate this, we re-partition our dataset as followings:

\paragraph{app generality}  Since we use multiple apps in weather domain and calendar domain, we use the data from one APP as the test set, and the other data forms the training set.

\paragraph{domain generality} We use the data from one domain as the test set, and the other data forms the training set.

We evaluate the performance of m-BASH on these datasets. The results are shown in Table \ref{tab:generality}. We can find that our system can still obtain a reasonable performance, and the results of app generality experiments are even comparable to the main experiment results of LayoutLM. This result shows that common operation logic does exist in APPs, and our system can gain a general comprehension of GUI operations. It is easily applied to a new app or a new domain without modification, which shows the effectiveness and potential of our system.

\begin{table}[h]
\centering
\begin{tabular}{lcc}
\toprule
\begin{tabular}[c]{@{}l@{}}Data Domain \\ of Test Set\end{tabular} & \begin{tabular}[c]{@{}c@{}}Action \\ Completion\\ Rate (\%)\end{tabular} & \begin{tabular}[c]{@{}c@{}}Turn \\ Completion\\ Rate (\%)\end{tabular} \\
\midrule
\textit{app generality} \\
an app of weather                                                  & 56.45                                                                         & 45.71                                                                       \\
an app of calendar                                                 & 69.84                                                                    & 23.17                                                                  \\
\midrule
\textit{domain generality} \\
weather                                                            & 41.96                                                                    & 21.04                                                                  \\
calendar                                                           & 62.39                                                                    & 19.20                                                                  \\
search                                                             & 59.40                                                                    & 16.24                                                                  \\
taxi                                                               & 37.68                                                                    & 21.72                                                                       \\
restaurant                                                         & 30.26                                                                    & 15.42                                                                  \\
hotel                                                              & 31.24                                                                    & 16.26                                                                  \\
\bottomrule
\end{tabular}
\caption{The results of generality experiments.}
\label{tab:generality}
\end{table}

\section{Related Work}
\subsection{Natural Language Commands on GUI}
Executing natural language commands on GUI is getting research interests recently. Some studies focused on semantic parsing~\cite{mazumder2021flin,pasupat2018mapping,xu2021grounding}, whose task is mapping the natural language query to the operations on websites. 
%PIXELHELP~\cite{li2020mapping} is a dataset for natural language commands on Android and the queries are collected from Pixel Phone Help pages. These studies only contain single-turn queries, while our proposed system is under the setting of multi-turn dialogue, which requires a comprehension of context.
%MiniWoB~\cite{shi2017world} is a reinforcement learning platform containing web-based GUI tasks, agents should finish different tasks described by a line of natural language by using the mouse and keyboard. The tasks in MiniWoB are handcrafted web pages and relatively simple, which makes it hard to apply on real APPs. 
Google Duplex~\citep{duplex} can operate websites to finish tasks like booking movie tickets or making restaurant reservations. However, it only supports limited websites and it's more a unified interface than a general dialogue system with GUI operating ability. Our proposed dataset contains real-world APPs and aims at training models with general GUI understanding.

\subsection{Programming by Demonstration on GUI}
Programming by Demonstration (PbD) systems focus on learning GUI tasks from human demonstration~\cite{riva2021etna,li2021glider,li2018kite,li2019pumice}. SUGILITE~\cite{li2017sugilite} records user's operations on GUI and generates a script for the learned task. APPINITE~\cite{li2018appinite} proposed to add descriptions for ambitious actions to enhance the robustness of the generated script. These systems generate scripts based on handcrafted rules and XML analysis, which is sensitive to GUI changes and exceptions. %\citet{sereshkeh2020vasta} proposed VASTA, a PbD system that leverages object detection and OCR to locate GUI components and use clustering algorithm to analyze user utterance. It can deal with more flexible user queries and GUI changes to some extent. However, it still relies on generating low-level task scripts rather than building comprehensive GUI semantics. 
In this work, we aim to build a robot that can work with general mobile GUI, rather than repeating operations. 
% We believe that a robot with a general understanding of GUI would also be able to gain better performance on PbD tasks.

\subsection{Visual Dialogue}

More and more researchers combine CV and NLP into the dialogue system and are involved inß a more challenging task, visual dialogue\cite{le2020video,agarwal2020history,le2020bist}. It can be seen as a multi-step reasoning process over a series of questions~\cite{gan2019multi}. \citet{gan2019multi} updated the semantic representation of the question based on the image and dialogue history. \citet{wang2020vd} proposed VD-BERT, a simple yet effective framework of unified vision-dialog Transformer that leverages the pre-trained BERT language models for Visual Dialog tasks. 
%Compared with image-grounded dialogue systems, video-grounded systems are more interesting and more challenging. \cite{le2020video} applied GPT-2 model to fuse multi-modal information over different levels. \cite{le2019multimodal} built a multi-modal transformer network to incorporate information from different modalities and further applied a query-aware attention to extract context-related features from non-text modalities. 
Visual dialogue focuses on understanding the image contents. Besides this, our tasks also require understanding the interactions between UIs.

\section{Conclusion}
In this paper, we proposed the task of GUI-based task-oriented dialogue system, which replaces the traditional TOD-specific API calls with GUI operations on real APPs. The advantage is that intelligent agents can perform tasks without the need of backend TOD-specific APIs and it doesn't rely on a domain-specific schema, which means it can be applied to a new domain easily. We collect META-GUI, a dataset with dialogues and GUI traces to serve as a benchmark. Our model shows promising results on the dataset, and we hope this work 
could stimulate more advanced methods on GUI-TOD. In the future, we will explore how to better incorporate GUI traces into our model and build the GUI semantics based on interactions.

\section*{Limitations}
We propose a GUI-based task-oriented dialogue system, which can perform GUI operations on real APPs to complete tasks. To verify the validity of the system, we collect META-GUI dataset, which contains dialogues and GUI operation traces. In real scenarios, an agent may not know how to complete the task presented by the user. In these cases, an agent might reply "It's too hard for me.", or something like this, which are not included in our dataset. In the future, we will augment the dataset to include such cases. Furthermore, the models we used are too large to be applied in mobile phones. It is important to compress the models, which we will attempt in the future.

\section*{Acknowledgments}
We sincerely thank the anonymous reviewers for their valuable comments. This work has been supported by the China NSFC Projects (No.62120106006, No.62106142), Shanghai Municipal Science and Technology Major Project (2021SHZDZX0102), CCF-Tencent Open Fund and Startup Fund for Youngman Research at SJTU (SFYR at SJTU).
% Entries for the entire Anthology, followed by custom entries
\bibliography{anthology}
\bibliographystyle{acl_natbib}

\clearpage

\appendix

\section{Details of Apps}
\label{app_data}

We list the information of applications used in Table \ref{tab:appdistribution}. To ensure the diversity of our dataset, we use 4 apps for weather domain, 3 apps for calendar domain, and 1 app each for the last 4 domains. We also list the number of turns belonging to each app. The total number of turns is larger than the actual number of turns, since that one turn may involve several Apps.

% Please add the following required packages to your document preamble:
% \usepackage{multirow}
\begin{table}[h]
\begin{tabular}{|l|l|l|}
\hline
Domain                    & Package                                                                            & \#Turn \\ \hline
\multirow{4}{*}{Weather}  & \begin{tabular}[c]{@{}l@{}}com.dailyforecast.\\ weather\end{tabular}               & 182    \\ \cline{2-3} 
                          & \begin{tabular}[c]{@{}l@{}}com.accurate.weather.\\ forecast.live\end{tabular}      & 291    \\ \cline{2-3} 
                          & \begin{tabular}[c]{@{}l@{}}com.graph.weather.\\ forecast.channel\end{tabular}      & 115    \\ \cline{2-3} 
                          & \begin{tabular}[c]{@{}l@{}}com.channel.weather.\\ forecast\end{tabular}            & 129    \\ \hline
\multirow{3}{*}{Calendar} & \begin{tabular}[c]{@{}l@{}}com.simplemobiletools.\\ calendar\end{tabular}          & 81     \\ \cline{2-3} 
                          & \begin{tabular}[c]{@{}l@{}}me.proton.android.\\ calendar\end{tabular}              & 777    \\ \cline{2-3} 
                          & \begin{tabular}[c]{@{}l@{}}com.google.android.\\ calendar\end{tabular}             & 52     \\ \hline
Search                    & \begin{tabular}[c]{@{}l@{}}com.google.android.\\ googlequicksearchbox\end{tabular} & 1616   \\ \hline
Taxi                      & com.ubercab                                                                        & 750    \\ \hline
Restaurant                & com.yelp.android                                                                          & 947    \\ \hline
Hotel                     & com.booking                                                                 & 942    \\ \hline
\end{tabular}
\caption{The information of Apps. The total number of turns is larger than the actual number of turns because some turns involve several APPs.}
\label{tab:appdistribution}
\end{table}

\section{Annotation System}
\label{annotation system}

\begin{figure}[h]
    \centering
    \includegraphics[width=0.45\textwidth]{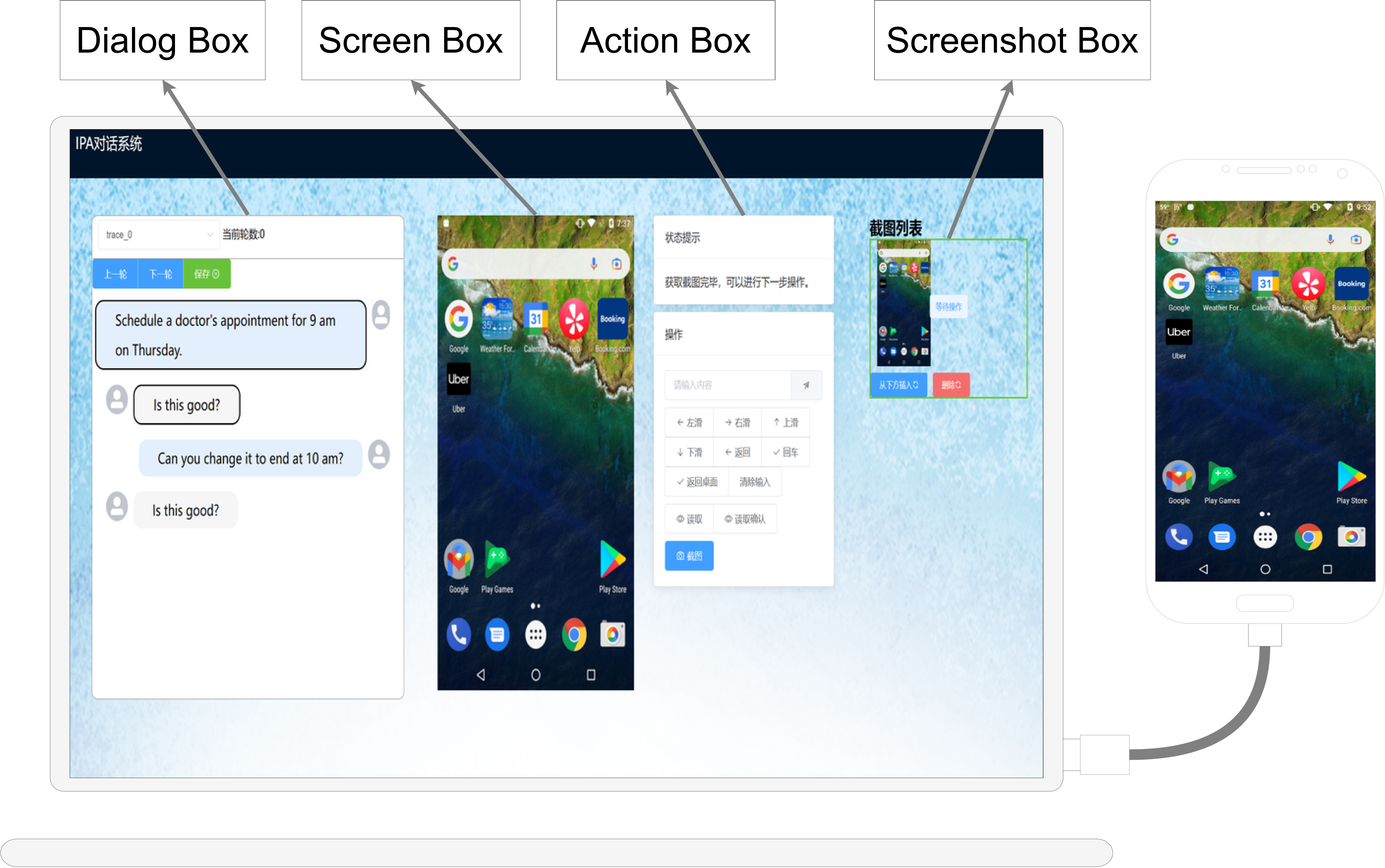}
    \caption{The illustration of our Annotation System.}
    \label{fig:annotation_system}
\end{figure}
The annotators can see dialogues in the Dialog Box and the current screen of smartphone in the Screen Box. Action Box proves buttons to control the smartphone, and the Screenshot Box records and displays the operation process.
% \newpage

\section{Example of View Hierarchy}
\label{view hierarchy}

\begin{figure}[h]
    \centering
    \includegraphics[width=0.485\textwidth]{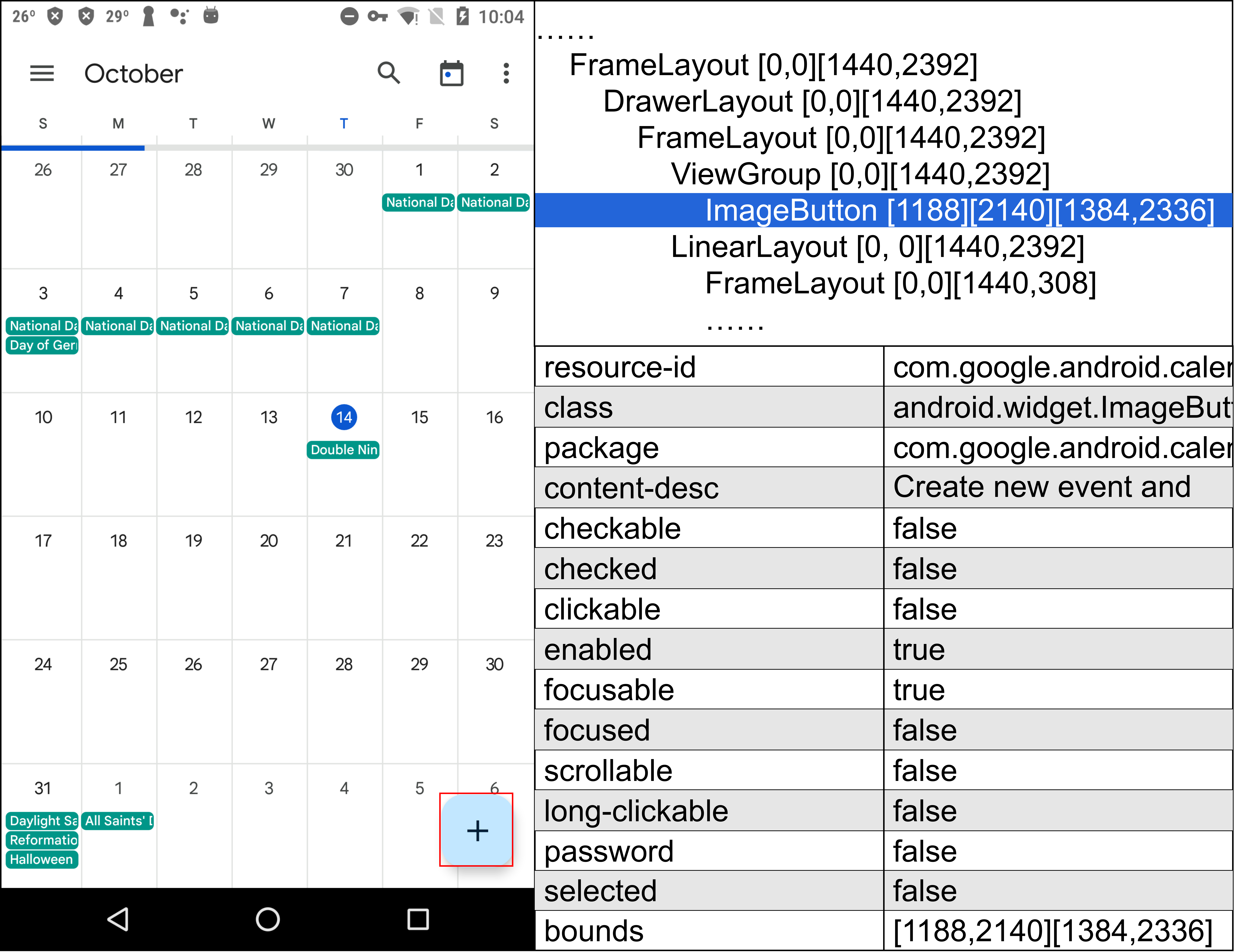}
    \caption{An example of the View Hierarchy for a given screen. The "+" button with a red border on the left-hand side corresponds to the highlighted element in the view hierarchy on the right-hand side.}
    \label{fig:viewhierarchy}
\end{figure}

\section{Data Review}
\label{datareview}

After annotation, we manually reviewed the data. The checklist includes: (1) whether the recorded GUI traces match the dialogues: we will check whether the GUI operations match the tasks proposed by the users, for example, whether the scheduled time is correct. (2) whether there are invalid operations due to the system error or misoperation: during annotation, some annotators may click a wrong position or swipe the screen mistakenly. The annotation system may sometimes run into failure. (3) whether there are redundant operations in the GUI trace: for example, some annotators may take screenshots of the same screen multiple times. 
% \section{Baseline}
% \label{baseline}
% In this section, we introduce the backbones we used in our experiments, BERT, LayoutLM and LayoutLMv2:
% \begin{itemize}
%     \item \textbf{BERT}~\cite{devlin2018bert} \ BERT is a pre-trained bidirectional language model. BERT is pre-trained with two tasks, Masked Language Model (MLM) and Next Sentence Prediction (NSP). It has shown great success in many NLP tasks. 
%     \item \textbf{LayoutLM}~\cite{xu2020layoutlm} \ LayoutLM is a variant of BERT. It jointly models interactions between text and layout information. It also leverage image features to incorporate visual information. LayoutLM achieves great results in several VDU tasks.
%     \item \textbf{LayoutLMv2}~\cite{xu2020layoutlmv2} \ LayoutLMv2 models the interaction among text, layout and image in a single multi-modal framework. It uses not only the existing masked visual-language modeling task but also the new text-image alignment and text-image matching tasks. Experiment results show that LayoutLMv2 outperforms LayoutLM by a large margin.
% \end{itemize}

\clearpage

\section{Example of the generated target during collecting}
\label{exampleoftarget}
\begin{figure}[h]
    \centering
    \includegraphics[width=0.48\textwidth]{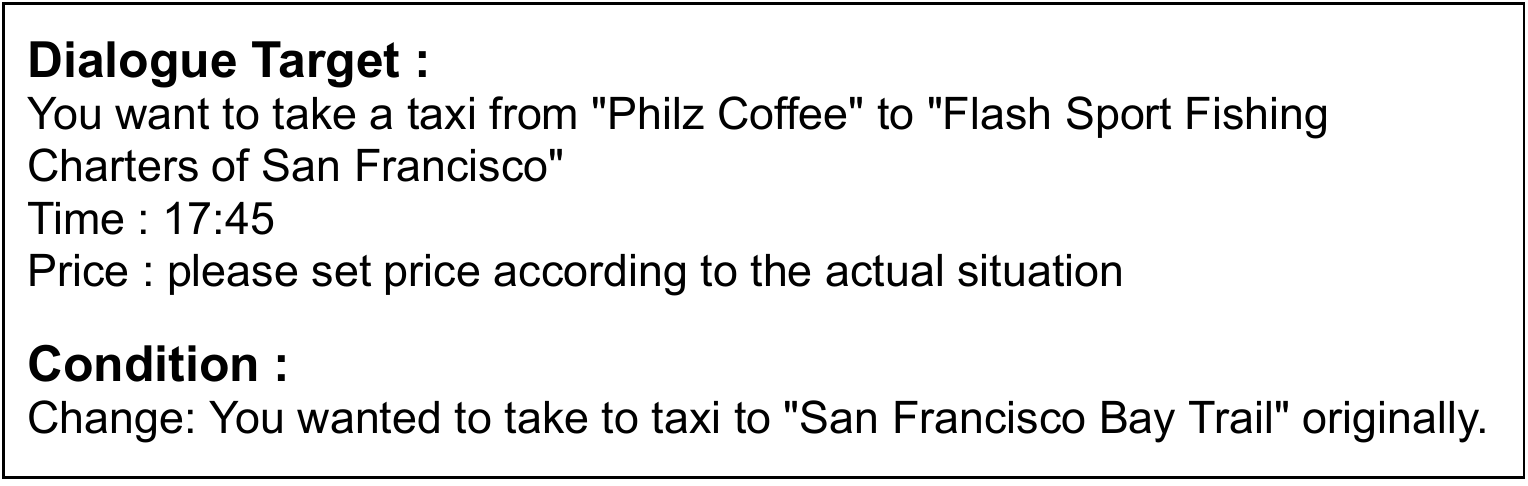}
    \caption{An example of the generated target.}
    \label{fig:dialog_target}
\end{figure}

\section{Examples of Item types}
\label{exampleofitemtypes}
We list an example for each of the item type in Table \ref{tab:examplesofitemtype}. There are 18 kinds of item types in total. And the corresponding items are highlighted with a red border.

\begin{table*}[]
    \centering
    \begin{tabular}{|c|c|c|c|c|c|}
    \hline
        Button & CheckBox & CheckedTextView & EditText & FrameLayout & Image\\
        \begin{minipage}{0.12\textwidth}
            \includegraphics[width=\textwidth]{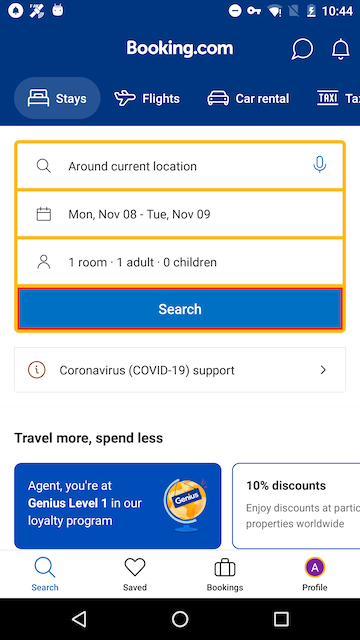}
        \end{minipage} & \begin{minipage}{0.12\textwidth}
            \includegraphics[width=\textwidth]{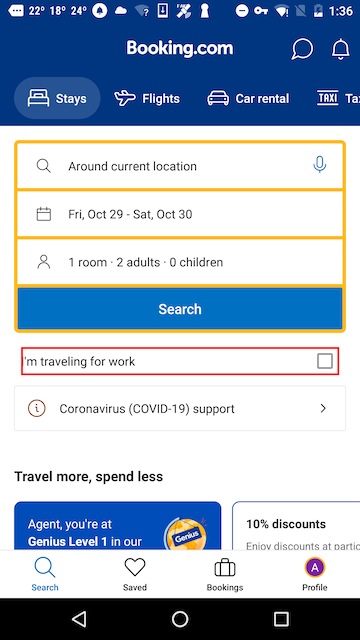}
        \end{minipage} & \begin{minipage}{0.12\textwidth}
            \includegraphics[width=\textwidth]{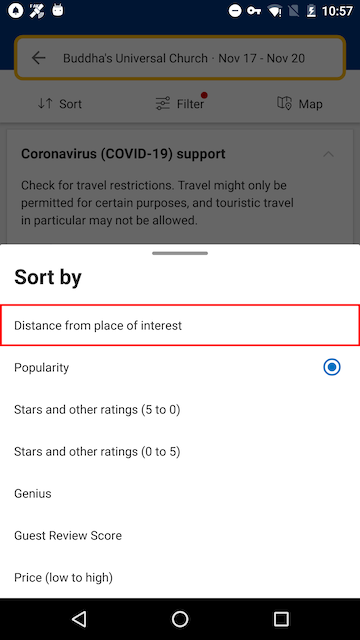}
        \end{minipage} & \begin{minipage}{0.12\textwidth}
            \includegraphics[width=\textwidth]{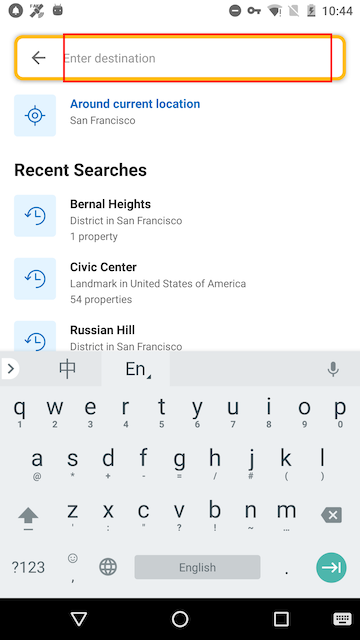}
        \end{minipage} & \begin{minipage}{0.12\textwidth}
            \includegraphics[width=\textwidth]{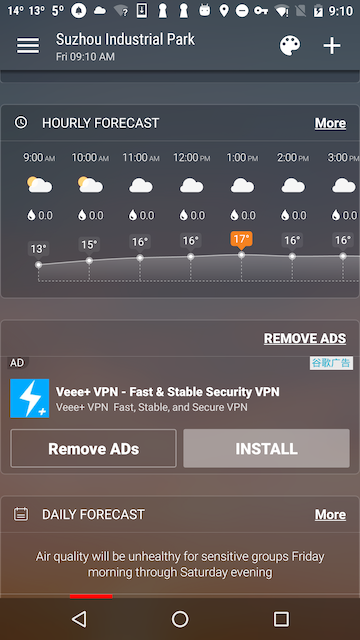}
        \end{minipage} & \begin{minipage}{0.12\textwidth}
            \includegraphics[width=\textwidth]{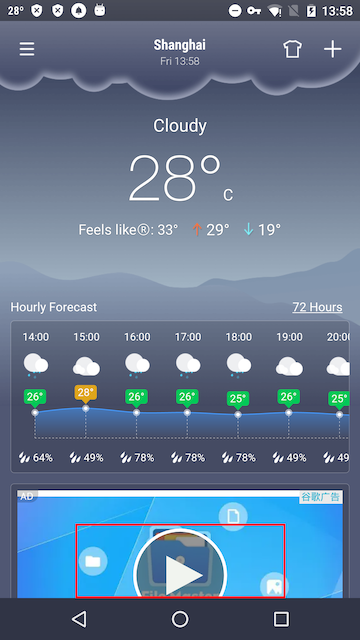}
        \end{minipage} \\ \hline
        ImageButton & ImageView & LinearLayout & ListView & RadioButton & RelativeLayout \\ 
        \begin{minipage}{0.12\textwidth}
            \includegraphics[width=\textwidth]{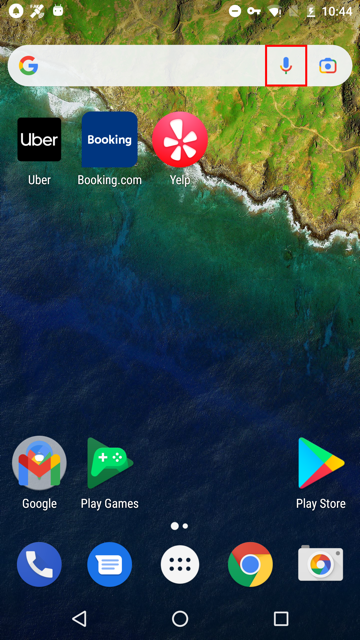}
        \end{minipage} & \begin{minipage}{0.12\textwidth}
            \includegraphics[width=\textwidth]{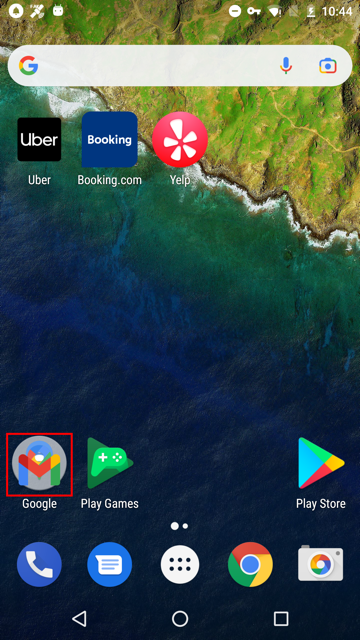}
        \end{minipage} & \begin{minipage}{0.12\textwidth}
            \includegraphics[width=\textwidth]{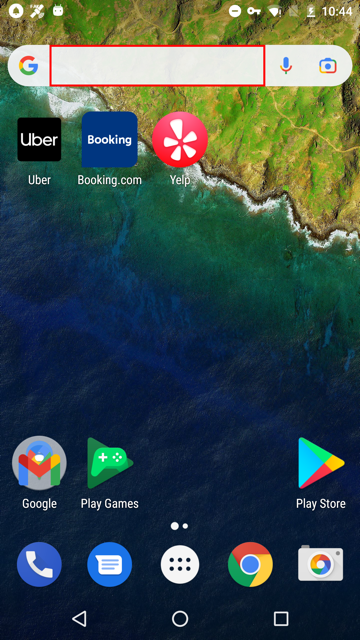}
        \end{minipage} & \begin{minipage}{0.12\textwidth}
            \includegraphics[width=\textwidth]{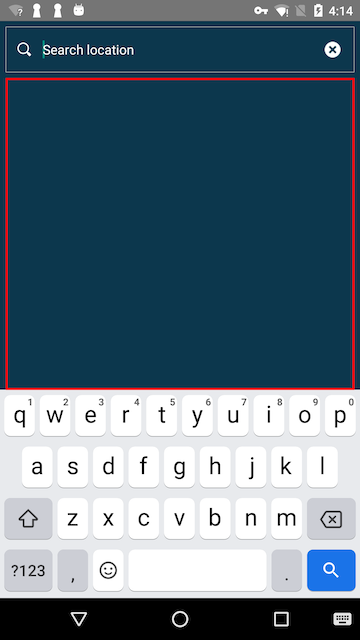}
        \end{minipage} & \begin{minipage}{0.12\textwidth}
            \includegraphics[width=\textwidth]{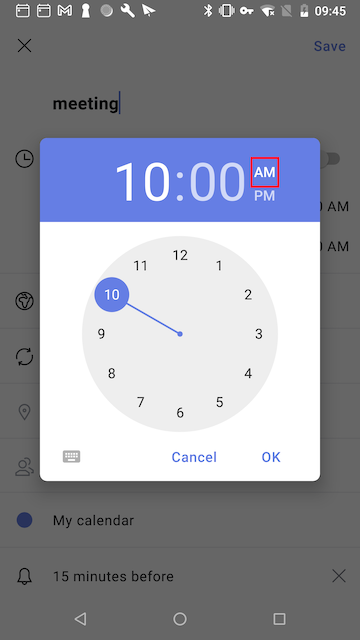}
        \end{minipage} & \begin{minipage}{0.12\textwidth}
            \includegraphics[width=\textwidth]{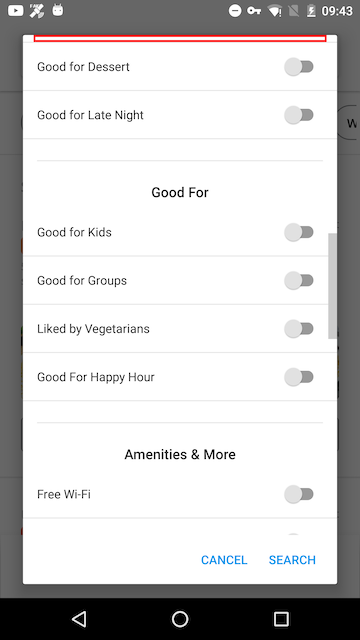}
        \end{minipage} \\ \hline
        Switch & TextView & ToggleButton & View & ViewGroup & WebView \\
        \begin{minipage}{0.12\textwidth}
            \includegraphics[width=\textwidth]{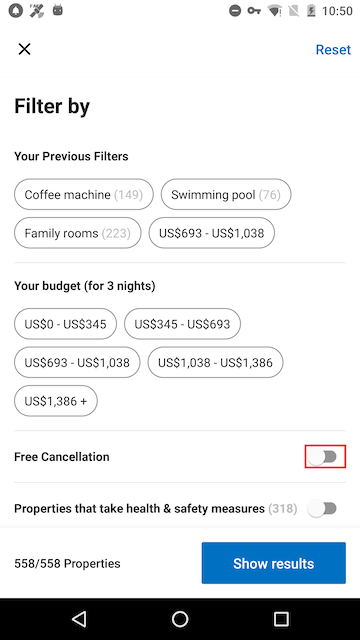}
        \end{minipage} & \begin{minipage}{0.12\textwidth}
            \includegraphics[width=\textwidth]{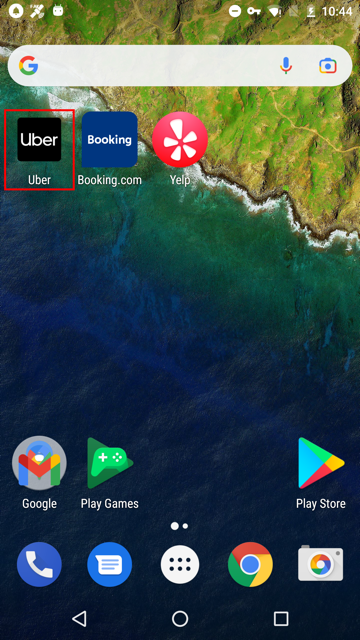}
        \end{minipage} & \begin{minipage}{0.12\textwidth}
            \includegraphics[width=\textwidth]{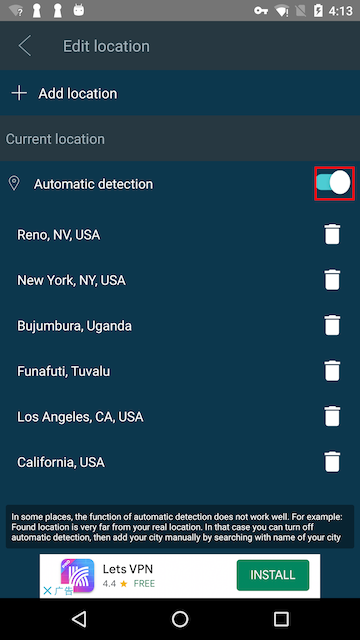}
        \end{minipage} & \begin{minipage}{0.12\textwidth}
            \includegraphics[width=\textwidth]{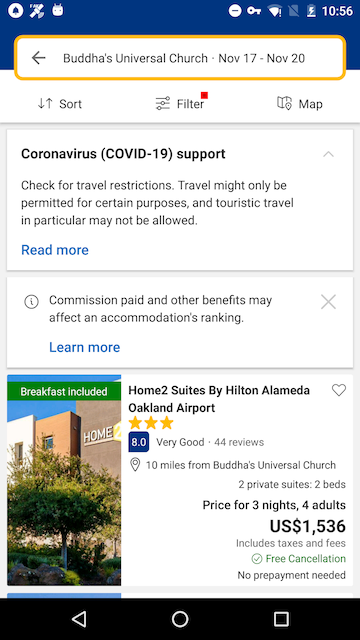}
        \end{minipage} & \begin{minipage}{0.12\textwidth}
            \includegraphics[width=\textwidth]{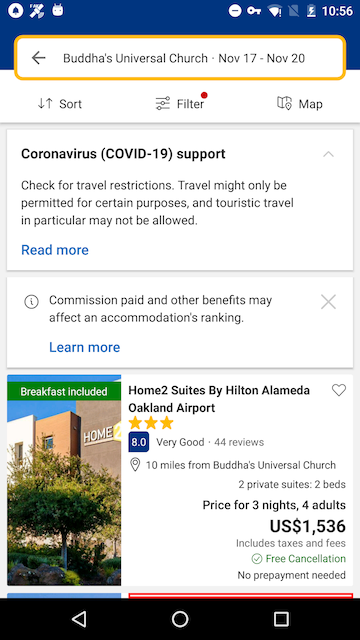}
        \end{minipage} & \begin{minipage}{0.12\textwidth}
            \includegraphics[width=\textwidth]{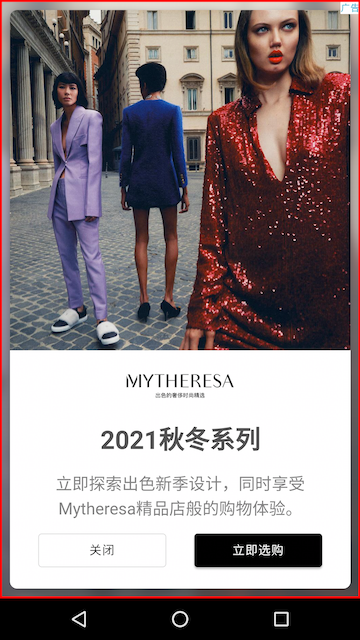}
        \end{minipage} \\ \hline
        
    \end{tabular}
    \caption{Examples of Item types.}
    \label{tab:examplesofitemtype}
\end{table*}

% \section{Heuristic Methods}
% \label{heuristicmethods}

% we use three heuristic methods in our experiments:

% \textbf{Random} We randomly predict action type and its corresponding parameters.

% \textbf{Frequency Method (FM)} We first calculate the frequency of each action type and its corresponding parameters. Then, we apply the results to the development set and generate the prediction according to the frequency.

% \textbf{Most Frequent Method (MFM)} Similar to the frequency method, we generate the prediction with the most frequent result.

% \section{Evaluation Metrics}
% \label{evaluationmetrics}

% We introduce accuracy, exact match, F1 score and BLEU score in this section.

% \textbf{Accuracy (Acc.)} This metric is used for evaluating whether the answers are successfully predicted. It is applied to the prediction of action type, target item and direction.

% \textbf{Exact Match (EM)} This metric is used to evaluate whether the text parameter of \texttt{Input} action is completely the same as the ground truth.

% \textbf{F1 score (F1)} This metric measures the overlap between the predicted answer and the ground truth. It is also applied to predicting the text parameter of \texttt{Input} action.

% \textbf{BLEU score} BLEU is a score for measuring the similarity of a candidate text to the references. We use it to evaluate the quality of the generated response texts.

\clearpage
\begin{figure*}[htbp]
    \centering
    \subfigure[Target Item prediction]{\includegraphics[width=0.32\textwidth]{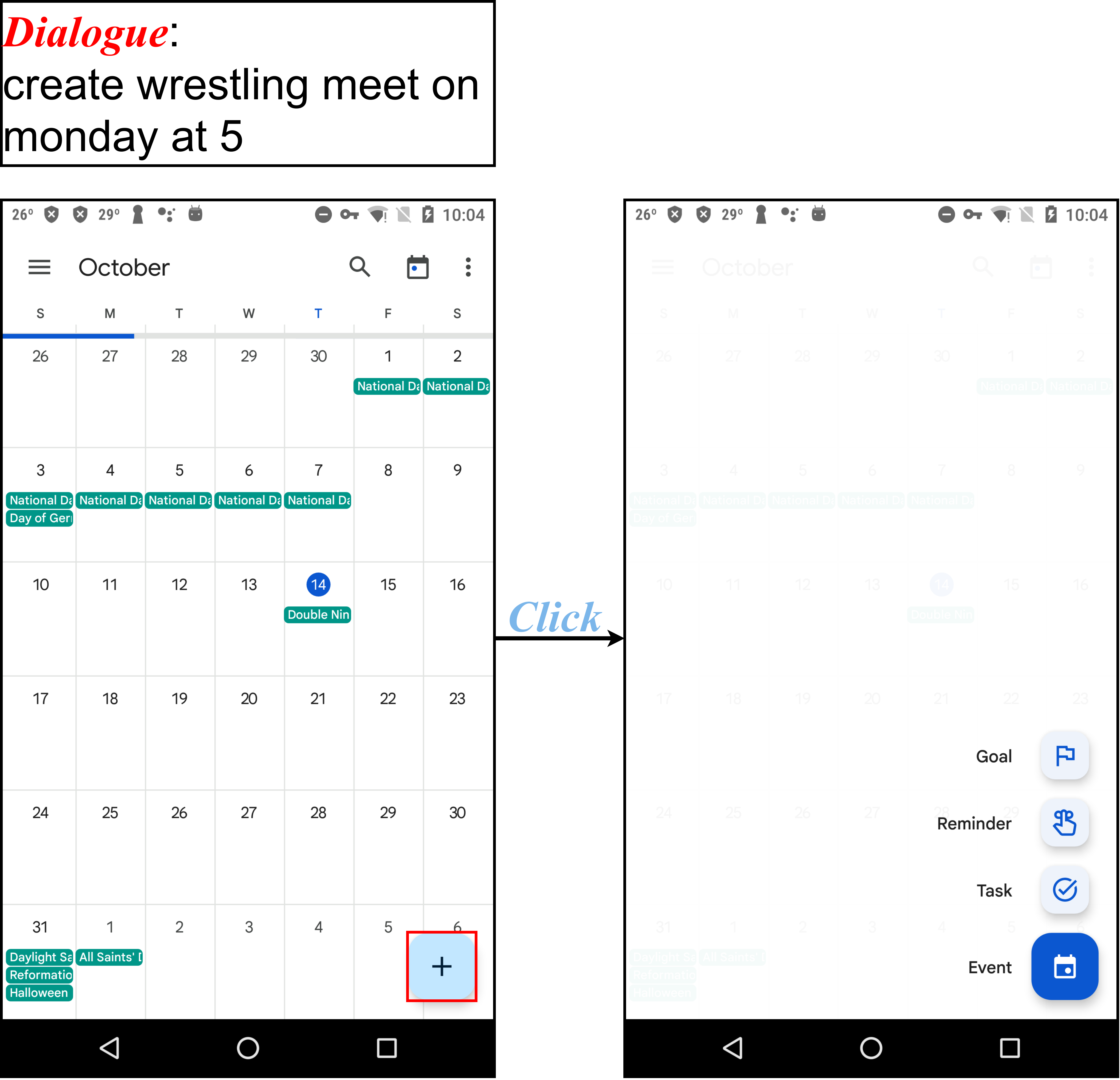} \label{fig:targetitemprediction}}
    \subfigure[Input Text prediction]{\includegraphics[width=0.32\textwidth]{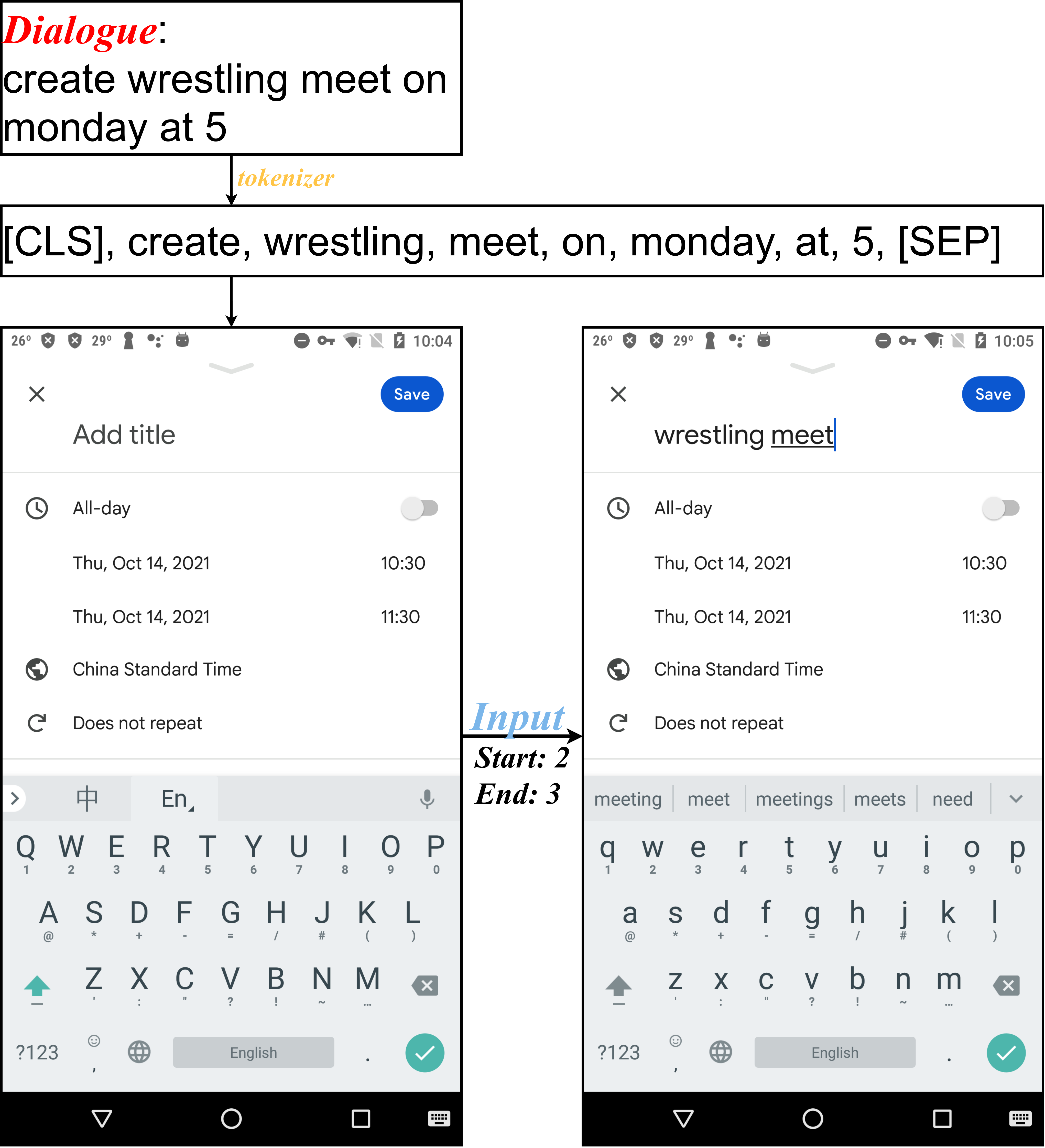} \label{fig:inputtextprediction}}
    \subfigure[Direction prediction]{\includegraphics[width=0.32\textwidth]{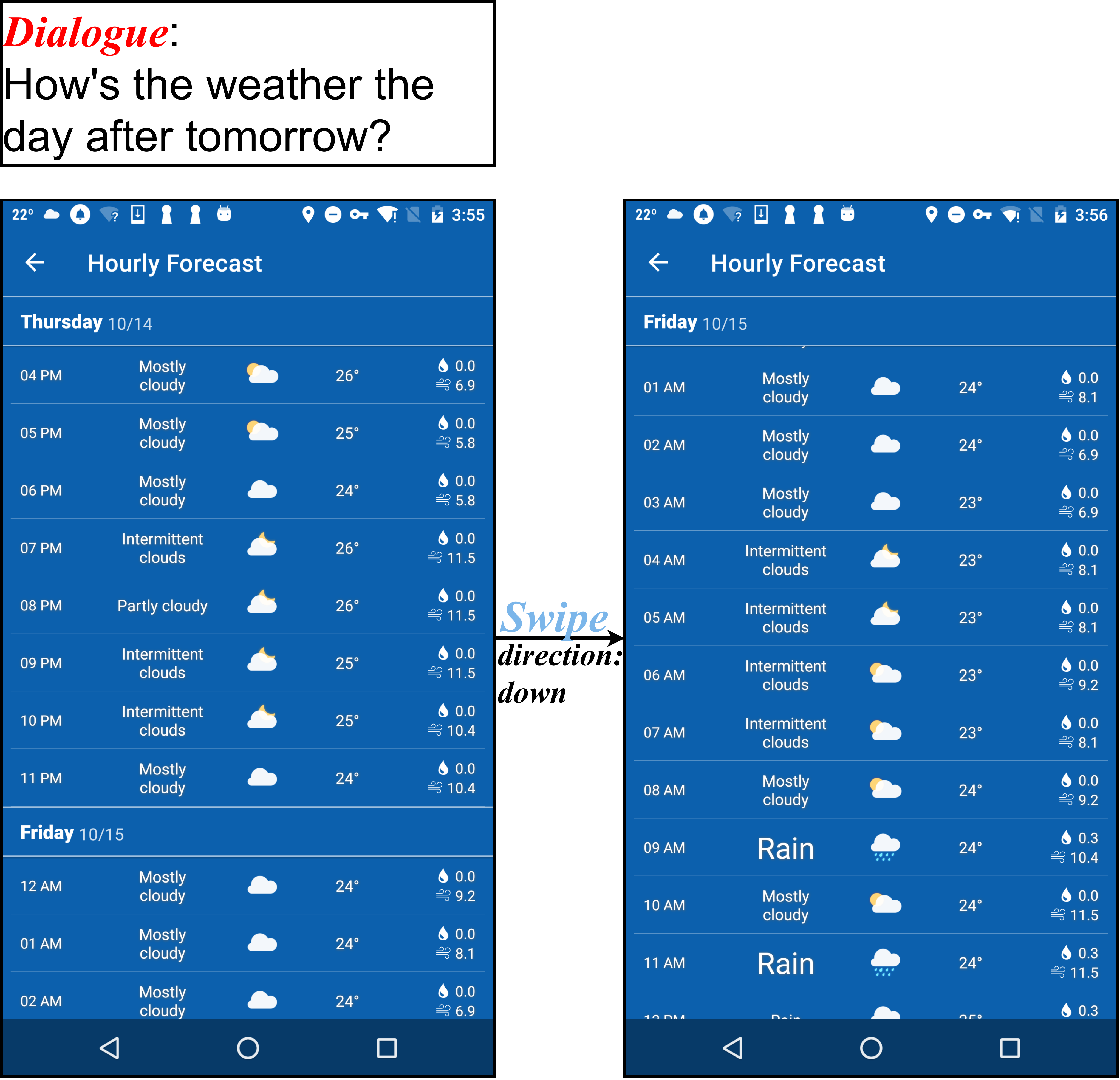} \label{fig:directionprediction}}
    
    \caption{Examples of parameter predictions.}
    \label{fig:parameterpredictions}
\end{figure*}

\section{Examples of parameter predictions}
\label{parameterpredictions}

We show some examples of parameter predictions in Figure \ref{fig:parameterpredictions}. Figure \ref{fig:targetitemprediction} shows an example of the prediction of target item. The left part shows the current screenshot, where the target item is highlighted with a red border. And the right part shows the screenshot after clicking the target item. Figure \ref{fig:inputtextprediction} shows an example of input text prediction. We first split the dialog into the token level, and then predict the text span. Figure \ref{fig:directionprediction} shows examples of direction prediction.
%\section{Add history information}
%\label{history}

%For action histories, we concatenate the most recent $H$ action types $\{t_{1:H}\}$ before the dialogue history as input:
%\begin{equation}
%    X = \{t_{1:H};w_{1:n};m_{1,1:l_1},\dots, m_{k,1:l_k}\},
%\end{equation}
%where $X$ stands for the input of encoder, $t$ represents the action type. 
%For screenshot histories, we use cross-attention mechanism to fuse histories. For each screenshot, we extract image features using the same method as image feature extraction module:
%\begin{equation}
%\begin{split}
%    \hat{\mathbf{V}}_{h+1} = \text{Attn}(\mathbf{W}_1&\mathbf{V}_{h+1},\mathbf{W}_2\hat{\mathbf{V}}_{h},\mathbf{W}_3\hat{\mathbf{V}}_{h}), \\
%    &1 \leq h \leq H-1,
%\end{split}
%\end{equation}
%where $\hat{\mathbf{V}}_{1} = \mathbf{V}_{1}$, $V_h$ stands for the image feature of $h$th screenshot, $V_H$ is the image feature of current screenshot, and $\mathbf{W}_1,\mathbf{W}_2,\mathbf{W}_3$ are trainable parameters. For the definition of Attn, we refer to~\cite{vaswani2017attention}.

\begin{figure*}[t]
    \centering
    \includegraphics[width=0.7\textwidth]{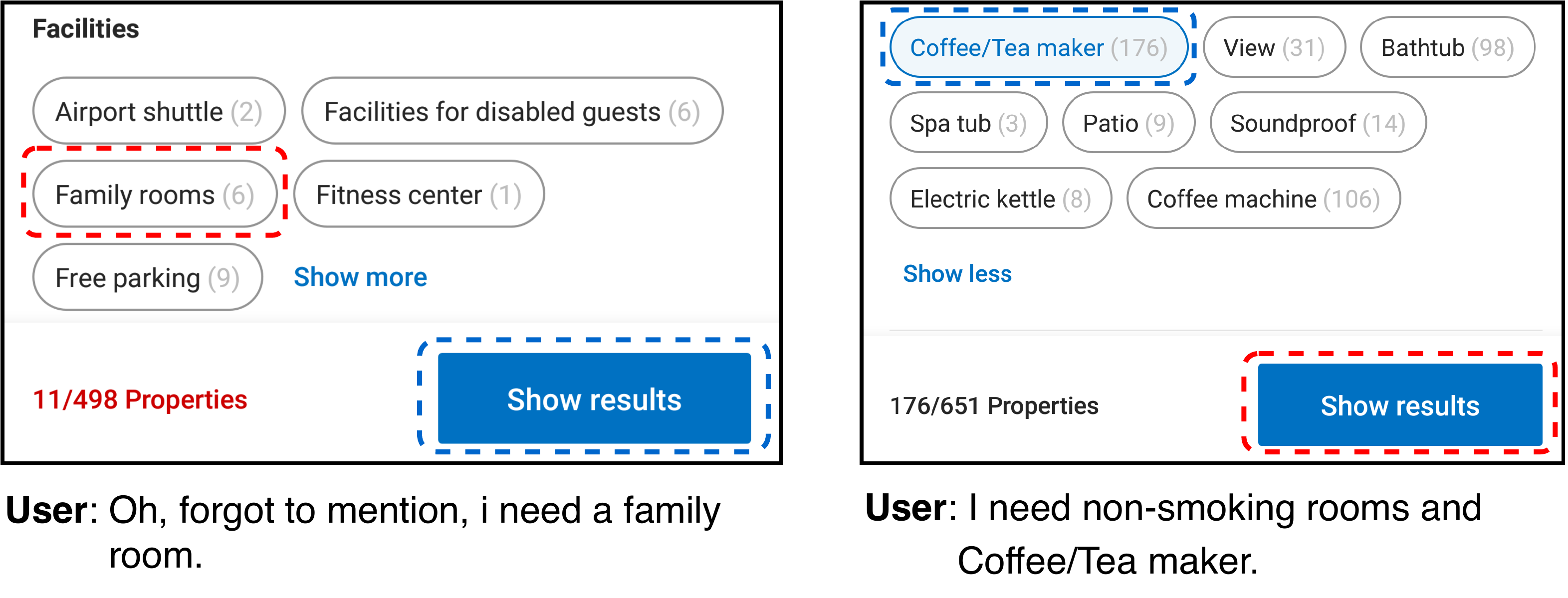}
    \caption{Case study. The predictions of m-BASH are marked by red boxes, which are the true answers, while the predictions of the BERT backbone model are marked by blue boxes.}
    \label{fig:casestudy}
\end{figure*}
\newpage
\section{Case study}
\label{casestudy}

To further show the importance of multi-modal information and history information, we select two samples, whose action type is \textit{click}, from our dataset and mark the predicted target items made by the BERT backbone model and m-BASH respectively. The result is shown in Figure \ref{fig:casestudy}. The predictions of m-BASH are marked by red boxes, which are the true answers, while the
predictions of the BERT backbone model are marked by blue boxes. It can be found that the reason why BERT backbone model makes mistakes is that it cannot distinguish whether the conditions are selected or not from text, which can be compensated by images and history information.

\end{document}